\documentclass[preprint,12pt,review,authoryear]{elsarticle}

\usepackage{amssymb}
\usepackage{amsmath}
\usepackage{booktabs}
\usepackage{url}
\usepackage{multirow}
\usepackage{longtable}
\usepackage{lscape}     % for landscape
\usepackage{makecell}
\usepackage{bbding}
\usepackage{threeparttable}
\usepackage{xcolor}
\usepackage[
    colorlinks=true,
    citecolor=blue,
    linkcolor=blue,
    urlcolor=blue
]{hyperref}

\journal{ISPRS Journal of Photogrammetry and Remote Sensing}

\begin{document}

\begin{frontmatter}



%\tnoteref{label1}}
%\tnotetext[label1]{}

\author[1]{Jinzhou Cao\fnref{equal}\corref{cor1}}
\ead{caojinzhou@sztu.edu.cn}
%\fntext[label2]{}
\cortext[cor1]{Corresponding authors}
\author[1]{Xiangxu Wang\fnref{equal}}
\fntext[equal]{Jinzhou Cao and Xiangxu Wang have contributed equally to the work.}
\author[1]{Jiashi Chen}
\author[2]{Wei Tu\corref{cor1}}
\ead{tuwei@szu.edu.cn}
\author[1]{Zhenhui Li}
\author[1]{Xindong Yang}
\author[1]{Tianhong Zhao}
\author[2]{Qingquan Li}

\address[1]{College of Big Data and Internet, Shenzhen Technology University, 
3002 Lantian Road, Pingshan District, 
Shenzhen 518118, Guangdong, China}

% \affiliation[1]{organization={College of Big Data and Internet, Shenzhen Technology University},
%           addressline={3002 Lantian Road, Pingshan District}, 
%           city={Shenzhen},
%            postcode={518118}, 
%            state={Guangdong},
%            country={China}}

\affiliation[2]{organization={Department of Urban Informatics, School of Architecture and Urban Planning, Shenzhen University},
           addressline={3688 Nanhai Avenue, Nanshan District}, 
           city={Shenzhen},
            postcode={518060}, 
            state={Guangdong},
            country={China}}

\title{Urban Representation Learning for Fine-grained Economic Mapping: A Semi-supervised Graph-based Approach} %% Article title

%% Abstract
\begin{abstract}
Fine-grained economic mapping through urban representation learning has emerged as a crucial tool for evidence-based economic decisions.
While existing methods primarily rely on supervised or unsupervised approaches, they often overlook semi-supervised learning in data-scarce scenarios and lack unified multi-task frameworks for comprehensive sectoral economic analysis.
To address these gaps, we propose SemiGTX, an explainable semi-supervised graph learning framework for sectoral economic mapping. 
The framework is designed with dedicated fusion encoding modules for various geospatial data modalities, seamlessly integrating them into a cohesive graph structure. It introduces a semi-information loss function that combines spatial self-supervision with locally masked supervised regression, enabling more informative and effective region representations. Through multi-task learning, SemiGTX concurrently maps GDP across primary, secondary, and tertiary sectors within a unified model.
Extensive experiments conducted in the Pearl River Delta region of China demonstrate the model's superior performance compared to existing methods, achieving R² scores of 0.93, 0.96, and 0.94 for the primary, secondary and tertiary sectors, respectively. Cross-regional experiments in Beijing and Chengdu further illustrate its generality. Systematic analysis reveals how different data modalities influence model predictions, enhancing explainability while providing valuable insights for regional development planning.
This representation learning framework advances regional economic monitoring through diverse urban data integration, providing a robust foundation for precise economic forecasting.

\end{abstract}


\begin{highlights}
\item A semi-supervised learning framework for socioeconomic mapping with balanced spatial self-supervision and direct supervision signals.
\item A multi-modal feature learning and graph-based fusion mechanism integrates diverse urban data sources at the regional representation level.
\item A multi-task framework for fine-grained sectoral GDP mapping with cross-sector interaction.
\end{highlights}

%% Keywords

\end{frontmatter}

% \linenumbers


\section{Introduction}\label{s1}

Regional economic mapping, particularly fine-grained GDP distribution across different geographical areas, forms the foundation for informed policy-making and balanced development, enabling responsive governance, and guides strategic decisions for regional economic advancement \citep{wu2021data, cao_UntanglingAssociationUrban_2024}.
Traditional economic estimation methods, including statistical analyses, econometric models, and spatial interpolation approaches \citep{victor2016autoregressive, stock2001vector, kim2018methods}, face notable limitations. These resource-intensive methods struggle with complex spatial heterogeneity, multicollinearity, and error propagation \citep{fan2021cpi}. The complex and non-linear nature of economic factors requires more sophisticated methodologies for precise spatial mapping.

The advent of the geospatial big data era has spurred the development of novel deep learning approaches for economic estimation using multimodal data. 
Nighttime light imagery correlates with economic activity and effectively captures the distribution of GDP in urban areas \citep{price2022global, shi2022population}, but underrepresents rural economies \citep{wang2024comprehensive, cao2025nighttime}. 
Similarly, remote sensing and street view imagery have been utilized \citep{li2022predicting, yong2024musecl}, yet often face limitations in spatial resolution or sampling bias in less developed regions. 
Complementary data sources, such as POIs and mobile positioning data, provide insights into facility distribution and human mobility patterns \citep{anand2022application}, offering valuable economic indicators for both urban and rural settings \citep{miao2021analyzing, pappalardo2015using}. Despite the richness of these diverse data sources, persistent challenges in spatial sampling bias, data scarcity \citep{bassi2024improving}, and effective modality integration continue to constrain their full analytical potential.

%In contrast, emerging geospatial vector data sources, such as Point of Interest (POI) data and mobile phone positioning data, provide different perspectives \citep{anand2022application}. These data reflect the distribution of facility development and human mobility patterns, offering more detailed insights into human-centric economic activities. POI data can display the distribution of businesses, service facilities, and other significant landmarks, which is crucial for understanding the economic ecosystems of both urban and rural areas \citep{miao2021analyzing}. Meanwhile, mobile phone positioning data can reveal residents' travel behaviors and social interactions, thereby providing richer contextual information for economic models \citep{pappalardo2015using}.

The persistent challenges of spatial sampling bias and data scarcity \citep{bassi2024improving} manifest as significant data imbalance and incompleteness, directly impairing the effectiveness of conventional supervised learning frameworks for urban economic prediction across diverse geographic regions \citep{richardson2021nowcasting, naaz2024forecasting}. Although unsupervised approaches offer alternative pathways by learning rich latent representations from unlabeled data, they typically struggle with spatial precision and exhibit sensitivity to data distribution shifts\citep{lee2023unsupervised}. 
Semi-supervised learning has emerged as a promising solution. Using unlabeled data strategically, through techniques such as self-training and consistency regularization\citep{yang2022survey}. By integrating labeled and unlabeled data, these approaches effectively reduce the spatial bias inherent in purely supervised frameworks while strengthening the representational capacity of unsupervised approaches \citep{jia2020semi, cai2024local}. 
This hybrid strategy has been validated in various deep learning frameworks, demonstrating robust performance on a variety of tasks \citep{xu2024deep, machicao2022deep}. However, a critical gap remains in the application of semi-supervised learning within multimodal graph-based models for urban economic forecasting tasks. Additionally, there is limited research on model interpretability, which is crucial for understanding the underlying mechanisms driving predictions and ensuring trust in decision-making processes. 

%Based on fundamental assumptions such as the smoothness assumption, cluster assumption, and the manifold assumption \citep{yang2022survey}, 
% Current models inadequately capture spatial dependencies in interconnected economies \citep{yu2022gdp}, while multi-source data integration demands sophisticated processing methods \citep{al2021review, jiang2023deep}.

%In addition, contemporary GDP mapping often overlooks sector-specific spatial patterns. Primary (agriculture, raw materials), secondary (manufacturing, construction), and tertiary (services) sectors each exhibit unique spatial distributions and responses to local conditions. For instance, while technological infrastructure primarily influences secondary sector distribution, population density predominantly affects tertiary sector patterns. This sector-specific variation significantly impacts mapping accuracy.

To address these challenges, we propose \textbf{SemiGTX}, an innovative graph-based semi-supervised graph learning framework. Our framework integrates diverse geospatial data modalities: street view images (SVI), point-of-interest (POI) and origin-destination (O-D) flows to comprehensively capture urban environments, social dynamics, and mobility patterns correlated with regional economic development. These heterogeneous data streams undergo specialized pre-encoding to generate modality-specific embeddings, which are then unified for comprehensive graph modeling. 
At the core of SemiGTX lies our proposed semi-info loss function, which elegantly balances self-supervised learning with masked supervised regression. The self-supervised component learns spatially interactive representations capturing complex geographical dependencies, while the supervised component leverages limited label information to guide accurate economic predictions.
We adopt a multi-task learning paradigm by decomposing GDP mapping into primary, secondary, and tertiary sector estimations for granular economic analysis, an area previously unexplored. Our comprehensive experiments demonstrate that SemiGTX achieves superior accuracy while maintaining robust performance across regions with varying data availability and quality, significantly advancing geospatial economic mapping.

The contributions of our study can be summarized as follows:
\begin{itemize}
    \item We develop SemiGTX, a novel framework for modality-specific encoding architectures and graph-based fusion mechanisms, enabling the comprehensive integration of heterogeneous urban data streams.
    \item We design an innovative semi-supervised loss mechanism that combines limited supervision with extended subgraph-level mutual information maximization, thereby improving the model's generalization and robustness for accurate sector-specific economic mapping.
    \item Attribution analysis techniques are integrated into the proposed framework to enable systematic analyses and visualization of the impact of multi-modal input features on model outputs, improving the explainability of the model. % and providing valuable guidance for regional development planning.
\end{itemize}

The structure of the paper follows:
Section \ref{s2} reviews related work on the mapping of economic indicators, graph-based urban representation learning and various learning modes applied to urban task. 
Section \ref{s3} details the proposed SemiGTX framework. 
Section \ref{s4} presents the study area, the data sources, the training and implementation of the model, together with comprehensive comparative experiments and evaluations. 
Section \ref{s5} provides an in-depth discussion of the model's heterogeneity, explainability, and limitations. 
Section \ref{s6} concludes the study.

%We develop SemiGTX, a novel semi-supervised graph learning framework that integrates spatial self-supervised learning of spatially-interacted representations with locally supervised regression through a novel semi-information loss function for accurate economic mapping.

%Extensive experiments demonstrate the superiority and versatility of the proposed SemiGTX model over existing methods.

\section{Related work}\label{s2}

\subsection{Economic indicators mapping}

Economic indicators serve as vital metrics of regional economic status and development, crucial for formulating strategic policies. 
Traditional approaches rely on community surveys for GDP and household income, which is resource-intensive and time-consuming \citep{custodio2023review}. 
Therefore, researchers have dedicated their efforts to developing data-driven forecasting methods using various data sources and approaches \citep{provost2013data}. Early forecasting methodologies employed classical statistical approaches, leveraging historical economic data to project future patterns. Notable examples include ARIMA for export forecasts \citep{victor2016autoregressive}, vector autoregressions (VARs) for unemployment and inflation forecasting \citep{stock2001vector}, and MIDAS for GDP forecasting \citep{kim2018methods}.

The field later witnessed a paradigm shift with the emergence of machine learning techniques. The research by \citet{yoon2021forecasting} compared gradient boosting and random forest models for GDP growth forecasting, while \citet{ulker2019unemployment} explored support vector regression (SVR) for unemployment and GDP forecasts. The remarkable performance of recurrent neural networks (RNNs) has garnered prominence, with implementations of LSTM \citep{siami2018forecasting} and its variants like ARIMA-LSTM \citep{fan2021well}, bi-LSTM, and convolutional LSTM \citep{murugesan2022forecasting}. However, these time series models demonstrate limited efficacy due to their reliance on unimodal data and neglect of multi-source factors.

The convergence of urban geospatial data and advanced deep learning has catalyzed new research for mapping economic indicators through multimodal spatio-temporal data.
These datasets encompass POI information, vehicle trajectories, remote sensing data, street view imagery, and social media data, providing robust support for economic mapping tasks.
Initially, researchers used single-source data approaches. For example, 
\cite{shi2022population} examined GDP-population relationships on spatial scales using nighttime light sensing, while \cite{price2022global} enhanced economic indicator mapping by integrating night and daytime remote sensing through transfer learning. Research in street view imagery \cite{fan2023urban} revealed the potential of street-level visual information for socio-economic assessment. The field then advanced to multisource data fusion. 
\citet{cao2022machine} and \citet{puttanapong2022predicting} demonstrated the effectiveness of combining diverse geospatial data in economic indicators forecasting. 

Despite these advances, existing methods typically focus on mapping single economic indicators or limited indicator sets (see Table~\ref{tab:works}), failing to capture economic heterogeneity across diverse industrial structures. For instance, mapping GDP across primary, secondary, and tertiary sectors requires modeling distinct dynamics, yet current methods often treat them as homogeneous targets.

\subsection{Graph-based urban representation learning}

Urban representation learning has emerged as a powerful paradigm for extracting meaningful information from urban data without arduous feature engineering. This approach transforms complex urban characteristics into numerical expressions while preserving spatial and semantic relationships. Recent years have witnessed various urban challenges, from crime prediction \citep{wang2017region, zhang2021multi} and land cover classification \citep{yao2018representing, luo2022urban} to socioeconomic indicator prediction \citep{wang2018urban, cui2018survey}. Early approaches used matrix factorization \citep{roweis2000nonlinear, belkin2001laplacian}, while later methods adopted word embeddings using random walks \citep{perozzi2014deepwalk, grover2016node2vec} and neighborhood sampling \citep{tang2015line}.
Graph Neural Networks (GNNs) \citep{abu2018watch, wu2022multi} represent a paradigm shift in urban representation learning. By naturally modeling urban elements as nodes and their relationships as edges\citep{cao_ResolvingUrbanMobility_2021}, GNNs enable more sophisticated modeling of urban structures \citep{liu2022review}, establishing themselves as a cornerstone for urban computing tasks \citep{wang2024hyperGCN}.
GNNs excelled in urban applications, from building pattern classification \citep{yan2019graph} and scene understanding \citep{xu2022framework} to traffic forecasting \citep{wang2023adaptive, li2017diffusion, jin2023spatio} and POI recommendation \citep{chang2020learning, wang2021attentive}.

% Table~\ref{tab:works}
Architectural innovations like STGCN's \citep{yu2017spatio} adaptable propagation and GAT's \citep{velickovic2017graph} attention mechanisms have further advanced urban applications. Transportation models such as GraphWaveNet \citep{wu2019graph} and HyGCN \citep{wang2024hyperGCN} demonstrate superior performance in capturing urban dynamics. Meanwhile, STG4DUF \citep{cao_DisentanglingHourlyDynamics_2025} has been proposed to address emerging scenarios, including mixed dynamic urban function mining. Despite these advances, multimodal urban data remains a challenge for conventional graph models. Approaches that effectively integrate massive heterogeneous urban data sources remain scarce, limiting their applicability in complex urban representations.

\subsection{Learning mode on urban tasks}

Supervised learning, as a well-established and effective method in machine learning, has been widely applied across urban studies such as crime prediction \citep{wang2017region, zhang2021multi}, urban functional zone classification \citep{zhang2017hierarchical}, and socioeconomic indicator prediction \citep{wang2018urban, cui2018survey}. 
However, due to limitations in data availability and quality, the field has witnessed parallel advancements in unsupervised learning methodologies, which focus on learning intrinsic features for generalized representations applicable to diverse downstream urban tasks. For instance, \citet{luo2022urban} developed an encoder-decoder module to integrate multi-view graph information, while \citet{li2023urban} adopted a dual-contrastive learning mechanism, leveraging InfoCE loss to learn global urban features and triplet loss to capture inter-region correlations within specific regions.
Deep Graph Infomax \citep{velivckovic2018deep} pioneered contrastive learning by maximizing mutual information between global and local graph features, enabling pattern extraction from unlabeled urban data.
This unsupervised paradigm has proven especially valuable for urban representation learning, enabling researchers to uncover meaningful patterns and relationships from vast amounts of unlabeled urban data.

Compared to unsupervised methods that rely solely on unlabeled data, leveraging limited labeled data alongside abundant unlabeled data offers a more practical direction by combining the strengths of both modalities. This strategy is especially pertinent in urban contexts, where the acquisition of labeled data is often constrained by scarcity or high costs. Existing studies have successfully applied semi-supervised frameworks \citep{laine2016temporal, berthelot2019mixmatch} to urban tasks. For example, \citet{deng2023supervised} demonstrated the effectiveness of balancing labeled and unlabeled data using similarity measures among POI data for urban functional zone classification under data-scarce conditions. Similarly, \citet{ma2023confidence} utilized pseudo-labeling techniques in semi-supervised learning to generate high-confidence labels from remote sensing images for land cover classification tasks. These studies highlight the potential of semi-supervised learning in addressing data scarcity challenges. However, a critical gap persists in the effective integration of semi-supervised learning into graph-based learning models for urban economic forecasting tasks (see Table~\ref{tab:works}), particularly in leveraging limited labeled data to enhance model performance and adaptability to complex urban environments.

\begin{table}[ht]
    \centering
    \caption{Comparison between SemiGTX and other main related works.}
    \resizebox{\linewidth}{!}{
    \begin{tabular}{c|ccc|cc|cc|l}
    \toprule[1.0pt]
         \multirow{2}{*}{\textbf{Method}}  & \multicolumn{3}{c|}{\textbf{Learning Mode}} & \multicolumn{2}{c|}{\textbf{Modal}} & \multicolumn{2}{c|}{\textbf{Task}} & \multirow{2}{*}{\textbf{Domain(s)}}\\ 
         & Supervised & Self-supervised & Semi-supervised & Signal- & Multi- & Signal- & Multi- &\\\toprule[1.0pt]
         ARIMA \citep{victor2016autoregressive}
             & \Checkmark & & 
             & \Checkmark & 
             & \Checkmark & 
             &  Export forecasts\\
        MIDAS  \citep{kim2018methods}
            & \Checkmark & & 
            & \Checkmark & 
            & & \Checkmark
            & GDP forecasts \\
        LSTM \citep{fan2021well}
            & \Checkmark & & 
            & \Checkmark & 
            & & \Checkmark 
            & Various economic indicators \\ \hline
        Visual Regression \citep{shi2022population}
            & \Checkmark & & 
            &  & \Checkmark
            & \Checkmark & 
            & Population forecasts\\
        Visual Transfer \citep{price2022global}
            & \Checkmark & & 
            & \Checkmark &  
            &  & \Checkmark
            & Economic mapping\\
        Street View Visual \citep{fan2023urban}
            & \Checkmark & & 
            & \Checkmark &  
            & \Checkmark & 
            &  Crime \& Health behaviors\\ \hline
        Autoencoder \citep{luo2022urban}
            & & \Checkmark  & 
            & & \Checkmark 
            & \Checkmark & 
            &  Urban representations\\ 
        Joint Learning \citep{zhang2021multi}
            & & \Checkmark  & 
            & & \Checkmark 
            & \Checkmark & 
            &  Crime prediction\\
        Triplet Loss \citep{li2023urban}
            & & \Checkmark  & 
            & & \Checkmark 
            & \Checkmark & 
            &  Region correlation \\ \hline
        SemiGTX (Ours)
            & \Checkmark & \Checkmark & \Checkmark
            & & \Checkmark 
            & & \Checkmark 
            & Fine-grained economic mapping \\ 
    \toprule[1.0pt]
    \end{tabular}
    }
    \begin{tablenotes}
		\item \textbf{Note}: SemiGTX supports three learning modes by adjusting the loss factor.
     \end{tablenotes}
    \label{tab:works}
\end{table}

\section{Methodology}\label{s3}

\subsection{Notations and definitions}

The essential notations used in this paper are summarized below (see Table \ref{tab:nota}), providing a clear reference for the mathematical expressions and algorithms presented in subsequent sections.

\begin{table}[t]
    \centering
    \caption{Summary of Notations}
    \resizebox{\linewidth}{!}{
    \begin{tabular}{cl}
    \toprule[1.0pt]
         Notation & Definition \\ \toprule[1.0pt]
         $\mathbf{X}_k^V$ & The physical environment feature vector of the $k$-th geographic unit. \\
         $\mathbf{X}_k^P$ & The POI feature vector of the $k$-th geographic unit. \\
         $\mathcal{G}$ & A graph modeled from the interactions between geographic units. \\
         $\mathbf{A}$ & Adjacency matrix of graph $\mathcal{G}$. \\
         $\mathbf{X}$ & Node features of graph $\mathcal{G}$. \\
         $\mathbf{E}$ & Edge features of graph $\mathcal{G}$. \\
         $\boldsymbol{h}_i$ & Output feature representation of positive samples at each layer of SemiGTX. \\
         $\boldsymbol{\tilde{h}}_i$ & Output feature representation of negative samples at each layer of SemiGTX. \\
         $\boldsymbol{s}_j$ & Feature representation of the $j$-th subgraph after dividing $\mathcal{G}$. \\
         $M$ & The number of nodes geographic units. \\
         $N$ & The number of subgraphs after dividing $\mathcal{G}$. \\
         $\mathcal{L}_{\text{info}}$ & Mutual information loss between $\boldsymbol{h}_i$, $\boldsymbol{\tilde{h}}_i$,  and $\boldsymbol{s}_j$, providing self-supervised signals.\\
         $\mathcal{L}_{\text{reg}}$ & Multi-task regression loss, providing supervised signals.\\
         $\lambda$ & Loss factor, used to balance $\mathcal{L}_{\text{info}}$ and $\mathcal{L}_{\text{reg}}$. \\
         $\mathcal{C}(\cdot)$ & Corruption function that generates negative samples from positive samples. \\
         $\text{MMPN}(\cdot)$ & Message-passing graph neural network. \\
         $\text{Attn}(\cdot)$ & Attention-like operation. \\
    \toprule[1.0pt]
    \end{tabular}
    }
    \label{tab:nota}
\end{table}

\subsection{Overview of SemiGTX}

\begin{figure}[t]
	\centering
	\includegraphics[width=\textwidth]{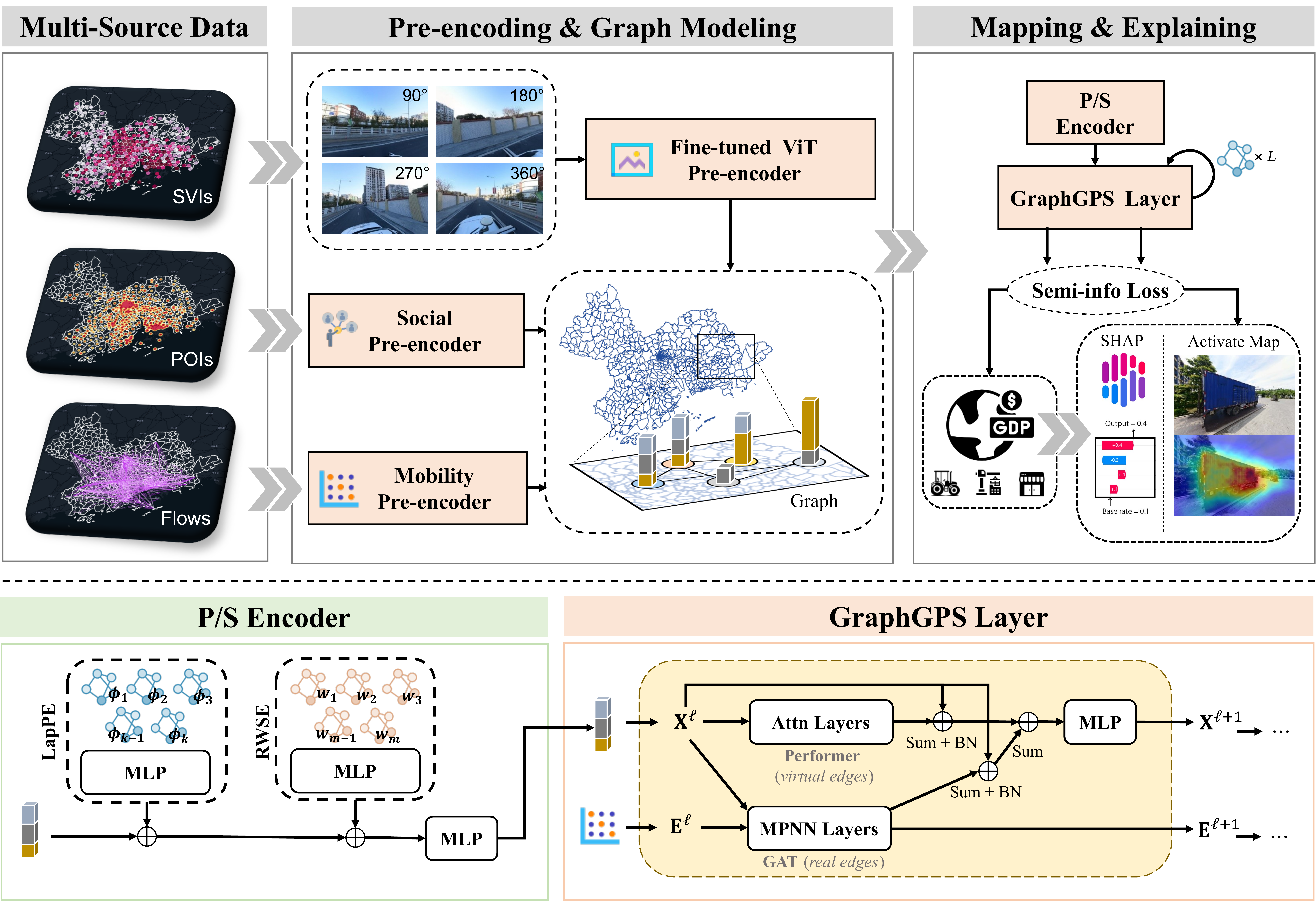}
    \caption{Overview of the SemiGTX Framework. Multiple data sources are integrated and transformed into a graph structure via specialized pre-encoders. Input data flows through P/S encoder and multiple GraphGPS layers, trained with semi-info loss, to generate multi-task outputs including GDP across primary, secondary, and tertiary economic sectors. SHAP-based interpretation techniques are also applied.}\label{fig:semigtx}
\end{figure}

The SemiGTX framework enhances multi-task economic indicator mapping in semi-supervised learning scenarios by strategically balancing self-supervision and direct supervision signals while processing multi-modal geospatial data.
As shown in Figure \ref{fig:semigtx}, SemiGTX integrates SVIs, POIs, and O-D flows to capture urban environments, social dynamics, and mobility patterns. 
These diverse data sources undergo modality-specific pre-encoders to generate embeddings that are then unified at the subdistrict level for comprehensive graph modeling. The Position or Structural (P/S) encoder, working alongside multiple GraphGPS layers, effectively captures complex interactions between different data modalities and spatial elements, synthesizing them into a holistic representation. 
To address limited labeled data in economic forecasting, SemiGTX employs a semi-info loss function that balances between regression loss and self-supervised loss through graph mutual information maximization. This approach enables robust spatial dependency modeling despite data sparsity by leveraging multi-task supervision with limited labeled data. For improved explainability, we implement a layer-wise trace-back strategy using SHAP analysis, providing valuable insights into the model's decision-making process and enhancing transparency in economic indicator predictions.

% Figure

\subsection{Pre-encoder design}\label{s3s2}

Geospatial data, inherently characterized by spatial interactions, is effectively modeled as a graph, where each node requires a well-defined representation to aggregate individual spatial data points into subdistrict-level units. 
Additionally, significant differences between SVIs, POIs, and O-D flows necessitate specialized encoders for each data type to enable seamless integration.

\subsubsection{Fine-tuned ViT pre-encoder}

For SVIs, inspired by large visual models such as the Segment Anything Model (SAM) \citep{kirillov2023segment}, a Vision Transformer (ViT)-based encoder is considered effective for transforming images into high-dimensional vectors. Therefore, in SemiGTX, a ViT-based encoder is utilized for SVIs.
Recognizing that existing pre-trained ViT models \citep{dosovitskiy2020image} are optimized for natural and general scenes rather than street views, as determined by the datasets used during their initial training, their encoding and representation capabilities for street view images would be weaker if directly applied to our task. Therefore, we design a custom autoencoder-based ViT pre-encoder specifically for SVI representation, training it from scratch to obtain fine-tuned weights tailored for urban street environments.

% Figure
\begin{figure}[t]
	\centering
    \includegraphics[width=\textwidth]{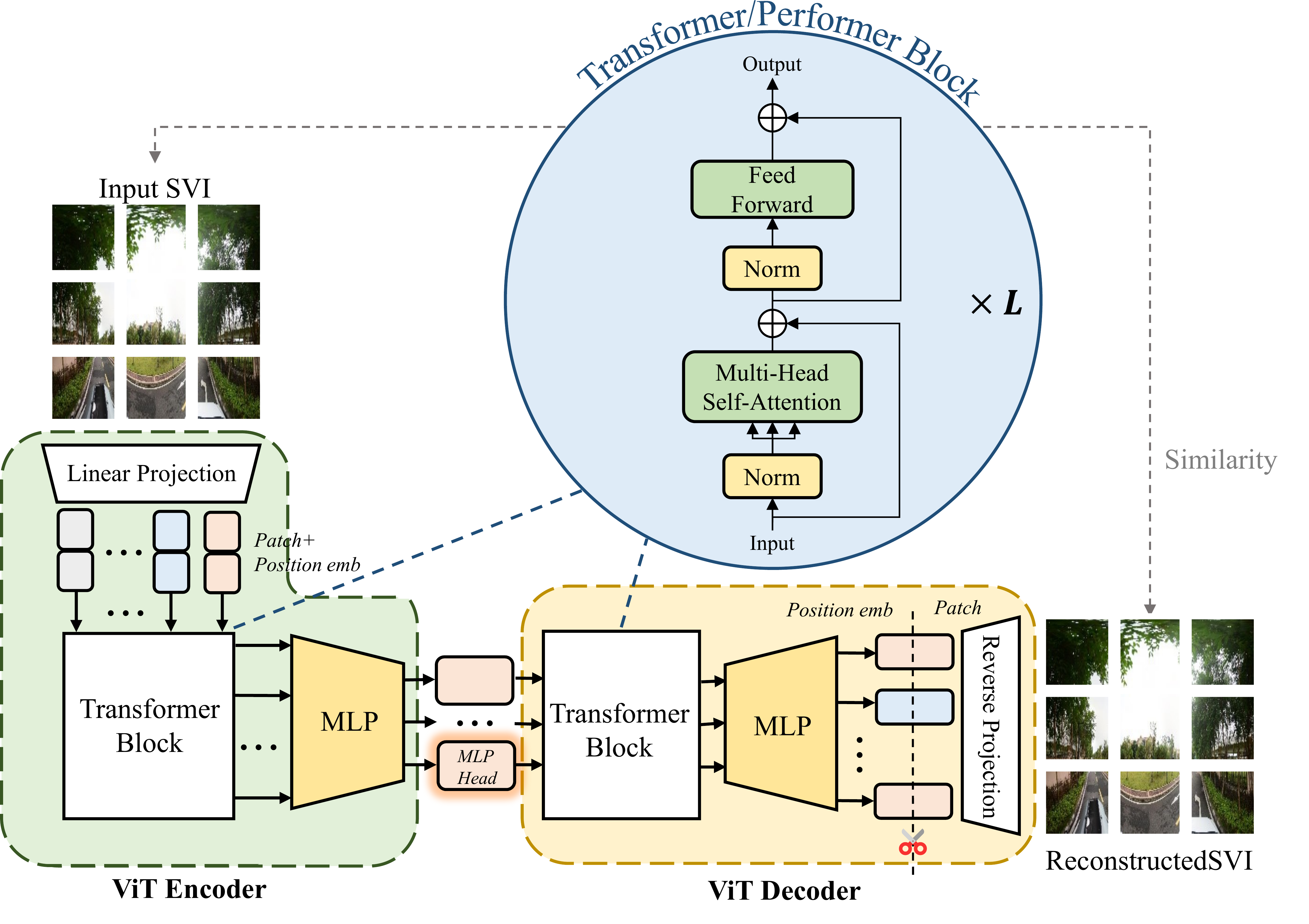}
    \caption{Structure of the fine-tuned ViT pre-encoder for SVIs. Each SVI is processed through the ViT encoder to generate its feature vector. During training, the ViT decoder reconstructs the image, and the model is trained by minimizing the discrepancy between the original and reconstructed images.}\label{fig:vit}
\end{figure}

Specifically, we modify the standard ViT architecture by truncating its final classification layer and add a ViT decoder that is symmetric to the ViT encoder module. As depicted in Figure \ref{fig:vit}, a street view image $\mathbf{I}_i$ is divided into equal-sized patches by the ViT encoder $\phi_v(\cdot)$ (including linear projection, transformer blocks, and MLP components). These patches undergo linear projections and are merged with position embeddings. Through attention mechanisms and MLPs, we obtain a representation matrix $\mathbf{X}^v_i=\left[\boldsymbol{x}^{v_i}_1, \boldsymbol{x}^{v_i}_2, \cdots, \boldsymbol{x}^{v_i}_p\right]\in\mathbb{R}^{d\times p}$, where each $\boldsymbol{x}^{v_i}_j\in\mathbb{R}^d$ represents the $j$-th $d$-dimesional latent vector of the $i$-th street view image $\mathbf{I}_i$, and $p$ denotes the total number of patches extracted from $\mathbf{I}_i$.
Utilizing the ViT decoder $\phi_v'(\cdot)$ (including transformer blocks, cutting and reverse projection), the matrix $\mathbf{X}^v_i$ is transformed into a reconstructed image $\mathbf{\tilde{\mathbf{I}}}_i$, such that $\mathbf{\tilde{\mathbf{ I}}}_i=\phi_v'(\phi_v(\mathbf{I}_i))$. 

Following autoencoder principles, we fine-tune the ViT weights by maximizing the similarity between the reconstructed and original SIVs.
The optimized weights are then frozen and used for pre-encoding SVIs. The \textit{MLP head} vector $\boldsymbol{x}^{v_i}_h$ defined as the first element of matrix $\mathbf{X}^v_i$(i.e., $\boldsymbol{x}^{v_i}_h:=\boldsymbol{x}^{v_i}_1$), will serve as the primary SVI embedding. All SVIs within the $k$-th subdistrict-level unit are aggregated using sum pooling to obtain the physical environment feature vector $\mathbf{X}_k^V$:
\begin{equation}
\centering
    \mathbf{X}_{k}^V=\frac{1}{|S_k|}\sum_{i\in S_k}\boldsymbol {x}_h^{v_i},
\end{equation}
where $S_k$ represents the index collection of points belonging to $k$-th unit and $|S_k|$ denotes the total number of points within this set.

\subsubsection{Social pre-encoder}

For POI data, which differ fundamentally from SVIs, a direct and effective approach is employed.
We computed the frequency of each POI category within each subdistrict-level unit, encoding POI data as categorical frequency vectors. The POI feature vector for the $k$-th unit,$\mathbf{X}_k^P$, is defined as:
\begin{equation}
\centering
    \mathbf{X}_k^P=\left(\sum_{i\in S_k}x_i^{c_1},\sum_{i\in S_k}x_i^{c_2},\cdots,\sum_{i\in S_k}x_i^{c_m}\right),
\end{equation}
where each binary variable $x_i^{c_j}\in\{0,1\}$ signifies whether the $i$-th POI belongs to category $j$, and $S_k$ represents the index set of POI points belonging to $k$-th unit.

\subsubsection{Mobility pre-encoder}

O-D flows naturally establish connections representing human mobility between two areas, serving as a crucial data source for constructing real edges in the graph. 
The mobility pre-encoder aggregates the raw point-to-point flow information into subdistrict-level units and constructs the adjacency matrix $\mathbf{A}$ and edge features $\mathbf{E}$ required for the graph structure.
Here, elements $a_{ij} \in \{0, 1\}$ in the adjacency matrix $\mathbf{A}$ indicate whether the $i$-th unit is connected to the $j$-th unit. The feature vector $\boldsymbol{e}_{ij}\in\mathbb{R}^2$ in $\mathbf{E}$ represents the edge features connecting the $i$-th unit to the $j$-th unit, determined by the frequency of the O-D records.

\subsubsection{Graph modeling}

In this study, an attributed weighted graph is defined as $\mathcal{G}=(V, \mathbf{A}, \mathbf{X}, \mathbf{E})$, where $\mathbf X$ and $\mathbf{E}$ represent the node and edge features, respectively. In SemiGTX, the node feature vector $\mathbf{X}_k$ for each node $k\in V$ (representing the $k$-th unit) is  created by concatenating SVI and POI embeddings:

\begin{equation}
    \mathbf{X}_k=\mathbf{X}_k^P \mathbin\Vert \mathbf{X}_k^V,
\end{equation}
where $\mathbf{X}_k^P$ and $\mathbf{X}_k^V$ represent the POI and SVI embeddings of the $k$-th unit, respectively.

As illustrated in Figure \ref{fig:semigtx}, the graph structure seamlessly integrates the encoding architectures tailored to each input data into a coherent and spatially structured representation. 

\subsection{Semi-supervised learning and task design}

\subsubsection{Position/Structural encoder}

Building on previous research, SemiGTX incorporates both positional and structural encoding to address the inherent limitations of Transformers in graph learning.
Our approach leverages the complementary strengths of Laplacian Positional Encoding (LapPE) and Random Walk Structural Encoding (RWSE) \citep{dwivedi2023benchmarking, dwivedi2021graph}. 
Specifically, letting a degree matrix $\mathbf{D}$ of the constructed graph be given, the Laplacian matrix $\mathbf{L}$ can be expressed as $\mathbf{L} = \mathbf{D} - \mathbf{A}$, where the eigenvectors and eigenvalues satisfy $\mathbf{L} \boldsymbol{\phi}_k = \lambda_k \boldsymbol{\phi}_k$.
\textbf{LapPE} computes the eigendecomposition of the Laplacian graph, selects the $K$ smallest eigenvalue-eigenvector pairs, and then uses an MLP to incorporate them as positional encodings in the form of node features. 
\textbf{RWSE} generates structural encodings by normalizing and linearly projecting the diagonal elements of the matrices derived from random walks of length 1 to $m$, represented as $\boldsymbol{w}_m=\text{diag}((\mathbf{D}^{-1}\mathbf{A})^m)$, which are also integrated as node features.
By integrating these encodings into the node representations (shown in Figure~\ref{fig:semigtx}), the model gains a more comprehensive understanding of the underlying spatial structures within the problem context.

\subsubsection{GraphGPS layer}
The GraphGPS layer processes the input graph through a combination of message-passing graph neural networks (MPNNs) and attention-based operations (Attns). 
Beyond handling node vectors connected by real edges established in the modeling phase, it also generates an attention matrix between nodes to create virtual edges as shown in Figure \ref{fig:semigtx}, thereby enhancing the representational capacity of the graph. The forward pass of the entire layer is defined as:

\begin{equation}
\begin{aligned}       
    \mathbf{X}_{\mathsf{MMPN}}^{\ell +1}, \mathbf{E}^{\ell +1}&=\text{MMPN}^{\ell}(\mathbf X^{\ell},\mathbf{E}^{\ell},\mathbf{A}),\\
    \mathbf{X}_{\mathsf{Tf}}^{\ell+1}&=\text{Attn}^{\ell}(\mathbf X^{\ell}),\\
    \mathbf{X}^{\ell+1}&=\text{MLP}^{\ell}(\mathbf{X}_{\mathsf{MMPN}}^{\ell+1}+\mathbf{X}_{\mathsf{Tf}}^{\ell+1}),
\end{aligned}\label{eq:gps}
\end{equation}
where MPNN$^{\ell}$ and Attn$^{\ell}$ are instances of an MPNN with edge features and a global attention mechanism at the $\ell$-th layer with their corresponding learnable parameters.

In SemiGTX, GAT \citep{velickovic2017graph} is chosen for the MPNN component and Performer \citep{choromanski2020rethinking} is selected for the Attn component. 
Performer is an efficient Transformer variant that uses the Fast Attention via Positive Orthogonal Random Features (FAVOR+) mechanism to accelerate attention computation while maintaining an architecture consistent with standard Transformers, as highlighted in the blue circular region of Figure \ref{fig:vit}.
GAT leverages attention mechanisms to dynamically weight the importance of neighboring nodes, making it particularly effective for complex graph structures. Formally, the node feature $\boldsymbol{h}_i$ is updated through a vanilla attention mechanism that aggregates information from its first-order neighbors $j\in N(i)$, resulting in the new representation $\boldsymbol{h}'_i$. The process is defined as follows:

\begin{equation}
\begin{aligned}
o_{ij} &= \mathrm{LeakyReLU}\left(\boldsymbol{a}^\top [W\boldsymbol{h}_i\;||\;W\boldsymbol{h}_j]\right), \\
\beta_{ij} &= \mathrm{softmax}(o_{ij})=\frac{\exp{(o_{ij})}}{\sum_{k\in N(i)}\exp{(o_{ik})}},\\
\boldsymbol{h}_i' &= \sigma\left(\sum_{j\in N(i)}\beta_{ij}W\boldsymbol{h}_j\right),
\end{aligned}\label{eq:gat}
\end{equation}
where $W$ is a learnable weight matrix, $\boldsymbol{a}$ is a learnable attention vector, $\beta_{ij}$ is is the importance assigned to $\boldsymbol{h}_j$ when updating $\boldsymbol{h}_i$, and $\sigma(\cdot)$ is an activation function.

\subsubsection{Semi-info loss}

\begin{figure}[!b]
    \centering
    \includegraphics[width=\textwidth]{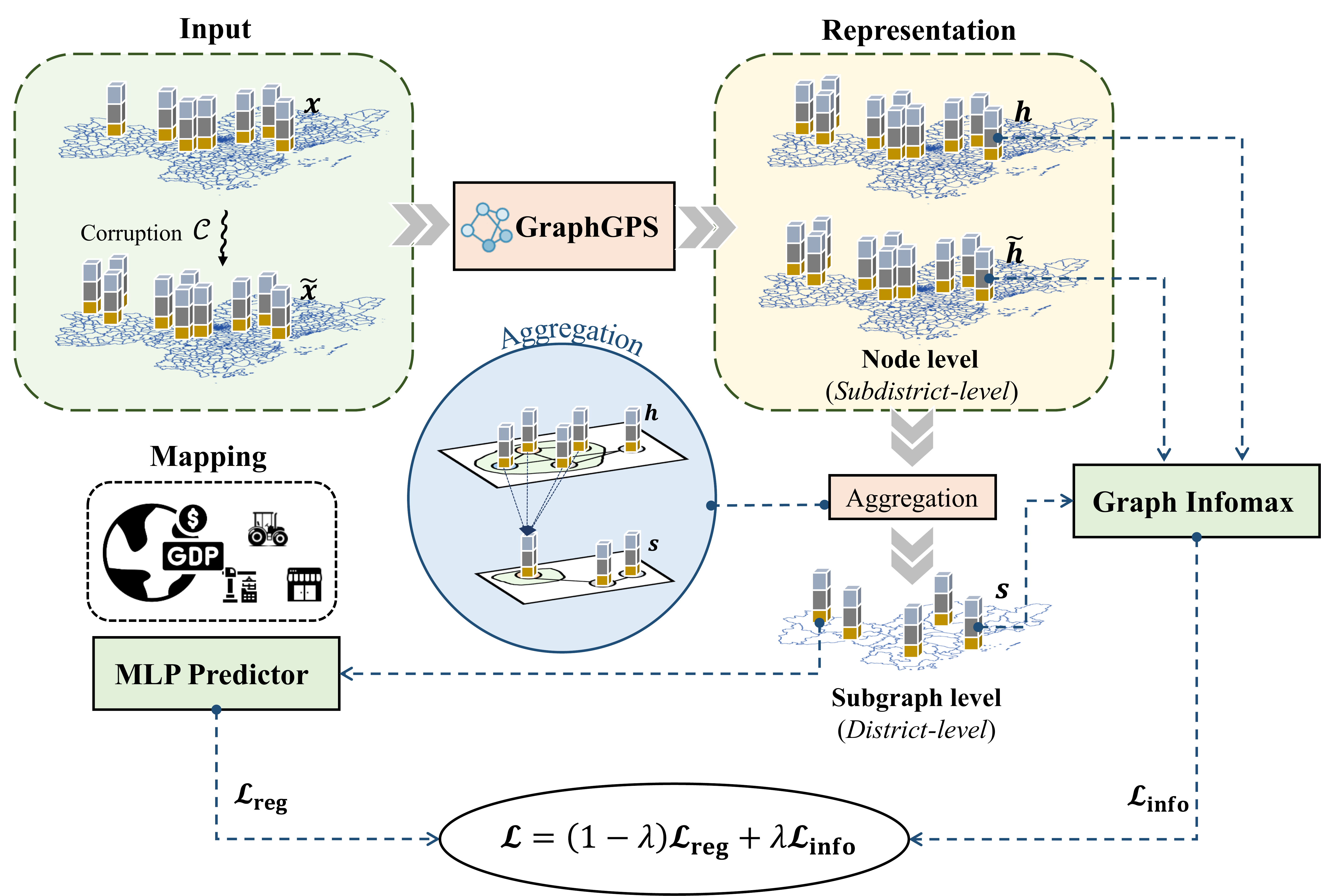}
    \caption{The workflow of the Semi-info loss. Positive sample features are transformed into negative samples via a corruption function. Subgraph-level representations are derived through aggregation and integrated with supervision signals via an MLP predictor to form regression loss. Additionally, positive and negative samples along with subgraph representations are used for self-supervised learning via the graph infomax method. The overall loss function balances these components using a weighting factor.}\label{fig:loss}
\end{figure}

For our semi-supervised scenario, we design a novel semi-info loss that balances the contributions of self-supervised learning based on spatial interactions with supervised learning based on local mapping. 
The self-supervised learning component extends Deep Graph Infomax (GIM) \citep{velivckovic2018deep} to enable subgraph-level (district-level), facilitating learning through contrast between positive and negative samples.
Specifically, our semi-information loss comprises three key steps, with the overall workflow illustrated in Figure \ref{fig:loss}.

\textbf{Corruption}: The constructed graph generates positive and negative samples through a corruption function $\mathcal{C}$. This function randomly removes edges and reorders nodes in the positive sample graph, creating negative samples $\boldsymbol{\tilde{x}}$ from positive samples $\boldsymbol{x}$. These samples then pass through multiple GraphGPS layers to obtain their respective node-level (subdistrict-level) representations $\boldsymbol{h}$ and $\boldsymbol{\tilde{h}}$.

\textbf{Aggregation}: The positive sample graph is partitioned into multiple subgraphs according to geographical districts. Each node is then aggregated according to its subgraph affiliation to form subgraph-level (district-level) representations $\boldsymbol{s}$, as shown in Figure \ref{fig:loss}. 

\textbf{Balancing}: The subgraph-level representations along with positive and negative node-level representations are jointly optimized by maximizing their mutual information $\mathcal{L}_{\text{info}}$, defined as: 
\begin{equation}
\begin{aligned}
    \mathcal{L}_{\text{info}}=&-\frac{1}{M}
    \bigg(\sum \limits _{j=1}^{N}\sum \limits _{i=1}^{M_i} \left[\log \mathcal{D}\left(\boldsymbol{h}_{i}, \boldsymbol{s}_{j}\right)\right]+\sum \limits_{j=1}^{N} \sum \limits _{i=1}^{M_i}\bigg[
    \log \left(1-\mathcal{D}\left(\widetilde{\boldsymbol{h}}_{i}, \boldsymbol{s}_{j}\right)\right)\bigg]\bigg),
\end{aligned}
    \label{eq:info}
\end{equation}
where $N$ is the number of subgraphs, $N'$ is the number of the unmasked districts, $M_i$ is the number of nodes associated with the $i$-th subgraph (with $\sum_{i=1}^{N}M_i=M$), and $M$ represents the total node count. The discriminator $\mathcal{D}(\boldsymbol{h}_i,\boldsymbol{s}_j)=\sigma(\boldsymbol{h}_i^{\top}\mathbf{W}\boldsymbol{s}_j)$ outputs a probability score indicating the alignment between the node representation $\boldsymbol{h}_{i}$ and the subgraph representation $\boldsymbol{s}_j$. When minimizing this loss, positive sample representations $\boldsymbol{h}_{i}$ become more aligned with their subgraph-level representations $\boldsymbol{s}_j$, while negative sample representations become increasingly differentiated.

In our transductive semi-supervised setting, ground truth (GT) values for certain districts are randomly masked. The model can only perform regression under the supervision signals from unmasked districts. These representations are processed through three prediction heads to generate sector-specific projections, which are then compared with GT values to compute the multi-task regression loss $\mathcal{L}_{\text{reg}}$:
\begin{equation}
    \mathcal{L}_{\text{reg}}=\frac{1}{L}\sum_{i=1}^{L}\frac{\alpha_i}{N'}\Vert\boldsymbol{\hat{y}}_{i}-\boldsymbol{y}_{i}\Vert^2,
    \label{eq:reg}
\end{equation}
where $L$ represents the number of economic indicators (in this study, $L=3$ corresponds to the value-added of three economic sectors), and $\alpha_i$ represents the weights assigned to different indicators. 
$\hat{\boldsymbol{y}}_i$ and $\boldsymbol{y}_i\in\mathbb{R}^{N}$ denote the predicted results and the GT values for the $i$-th indicator, respectively.

Ultimately, the two loss components are balanced through a weighting factor $\lambda\in[0,1]$, forming our comprehensive semi-info loss:
\begin{equation}
    \mathcal{L}=\lambda \mathcal{L}_{\text{info}}+(1-\lambda)\mathcal{L}_{\text{reg}}.
    \label{eq:loss}
\end{equation}

\section{Experiment and results}\label{s4}

\subsection{Study area}
The experimental study area encompasses the Pearl River Delta (PRD) region in Guangdong Province, a metropolitan cluster of nine major cities: Guangzhou, Shenzhen, Zhuhai, Foshan, Huizhou, Dongguan, Zhongshan, Jiangmen, and Zhaoqing (for specific spatial divisions, see \ref{app2}, Figure \ref{fig:division}). Situated in southern China, the PRD has emerged as one of the world's most vibrant megalopolises, distinguished by complex urban networks and diverse geographical features.
This region has become a center of technological innovation and economic growth, hosting numerous Fortune 500 companies and cutting-edge industrial parks. As China's primary manufacturing hub and a vital gateway to international trade, the economic importance of PRD extends worldwide, playing a key role in worldwide supply chains and technological development.

Our study adopts a multiscale framework aligned with China's administrative hierarchy. At its foundation, all multimodal data are standardized and integrated at the urban subdistrict level, representing the finest granularity of administrative units in China. 
These subdistrict-level features are then aggregated to the district level for economic indicator mapping, allowing microscopic and macroscopic analysis. 
%This hierarchical approach captures local patterns while revealing broader socioeconomic dynamics, which is critical for monitoring the PRD's rapid urbanization and economic transformation.

\subsection{Datasets}

\subsubsection{The GDP data}

Sectoral value-added data for PRD districts were sourced from the Guangdong Provincial Bureau of Statistics\footnote{\url{https://stats.gd.gov.cn}}. 
The data cover three economic sectors: primary, secondary and tertiary. 
The primary sector includes agriculture, forestry, animal husbandry and fishing activities, providing essential raw materials for the economy. The secondary sector consists of manufacturing, construction, and energy production, which transform these raw materials into finished goods and infrastructure. The tertiary sector encompasses service industries, including retail, education, healthcare, and financial services, which are integral to modern economic systems.

A notable administrative characteristic in the study area is that Dongguan and Zhongshan operate under a unique administrative structure, where they are directly subdivided into subdistrict-level units without an intermediate municipal district layer. To maintain analytical consistency, we treated these cities as equivalent to municipal districts in other PRD cities. The final dataset comprises value-added figures for the three economic sectors in 50 districts, with measurements denominated in 100 million Chinese yuan (see \ref{app2} Table \ref{tab:gdp} for details).

\subsubsection{The SVI data}

Street view images(SVIs) are photos taken from the street level, offering comprehensive visual representations of the urban environments, highlighting architectural styles, vegetation, and infrastructure.
We collect SVIs at the subdistrict level. For each sub-district, approximately 200 points were uniformly distributed along the road network. Using the Baidu Maps API\footnote{\url{https://lbsyun.baidu.com}}, images were retrieved at four viewing angles (90 °, 180 °, 270 ° and 360 °). In total, a total of 68,958 SVIs with a resolution of $480\times 320$ were collected.

\subsubsection{The POI data}
The study incorporates 3,960,271 POIs obtained from AMAP\footnote{\url{https://ditu.amap.com}}, China's leading digital mapping service. These POIs are systematically classified into 16 primary categories, including transportation facilities, leisure and entertainment venues, corporate enterprises, and medical facilities.
Further granularity is achieved through 140 medium category labels, which encompass specific designations such as three-star hotels, specialized hospitals, world heritage sites, and intermediary service providers. A spatial join operation was implemented to aggregate and analyze the distribution and frequency of POIs across all subdistricts. Detailed results are provided in \ref{app2} (see Table \ref{tab:poi} for an example visualization).

\subsubsection{Mobile phone positioning data}

The study uses anonymized mobile phone positioning data from China Unicom \footnote{\url{http://www.chinaunicom.com}}, a major national telecommunications operator. 
This dataset captures human mobility patterns through user movements between base station locations. To ensure strict privacy standards, the data underwent extensive aggregation at an elevated spatial scale. The analysis employed a 500-meter grid resolution to quantify movement volumes between the origin and destination cells, establishing an optimal compromise between analytical precision and privacy safeguards.

The dataset encompasses 34,960,199 hourly origin-destination movement records. To account for the weekly cyclical nature of human mobility, we computed average flow patterns for each day of the week and subsequently aggregated these patterns to derive annual mobility metrics. For subdistrict-level analysis, we transformed grid-based measurements into administrative unit metrics. Where grid cells spanned multiple subdistricts,  we distributed human flow volumes proportionally according to the respective area coverage ratios, ensuring accurate spatial representation of mobility patterns within administrative boundaries.

\subsection{Implementation details}

The fine-tuned ViT pre-encoder was initially trained to serve as the visual encoder, a pivotal component facilitating subsequent data fusion processes. 
The ViT encoder network was initialized with pre-trained weights from SAM to accelerate training, while the additional ViT decoder network was initialized using Xavier initialization.
For detailed parameter configurations, see \ref{app1}. 
After randomly selecting 40\% of the total images for training, the entire dataset is utilized for inference. Finally, we used the Structural Similarity Index (SSIM) \citep{wang2004image} as the evaluation metric to quantify image similarity, achieving an average SSIM value of 0.91, which falls within the favorable range of 0.80 to 1.00, indicating high visual similarity.  

For the graph learning model, the dimensions for both positional encoding and structural encoding were set to 16, while the output dimensions for MPNN and Attn were set to 512. The prediction head consisted of an MLP with 2 hidden layers activated by ReLU. 
The weights $\alpha$ for the multi-task regression losses were set to 0.1, 0.01, and 0.001 for the primary, secondary, and tertiary sectors, respectively, based on their relative scales.
All data were divided into training, validation, and test sets in a 7:1:2 ratio.
For additional details on the parameter settings, see Table \ref{tab:gps}.
After setting the learning rate to 9e-4 and completing the training in a Linux environment using a single NVIDIA A6000 GPU, the subdistrict-level representations were successfully generated.

\begin{table}[t]
    \centering
    \caption{The hyperparameter configurations of SemiGTX}
    \fontsize{10pt}{10pt}\selectfont
    \begin{tabular}{ccc}
    \toprule[1.0pt] %\hline
        \multirow{1}{*}{} & Parameter & Value \\ \toprule[1.0pt] %\hline
        \multirow{4}{*}{LapPE}  & PE dimension & 16\\
         &Embedding dimension & 512\\
         &Layers & 3\\
         &Max freqs & 10\\ \hline
         \multirow{4}{*}{RWSE} & PE dimension & 16\\
         &Embedding dimension & 512\\ 
         &Layers & 3\\
         &RW steps & 20\\ \hline
         \multirow{6}{*}{GPS} &Embedding dimension & 512\\ 
         & GNN type & GAT\\
         &Attention type & Performer\\
         &Attention heads & 8\\
         &Layers & 5\\
         &Activation & ReLU\\
    \toprule[1.0pt] %\hline
    \end{tabular}
    \label{tab:gps}
\end{table}

\subsection{Evaluation metrics}

To quantitatively assess our method's performance in the GDP mapping task, we employ multiple complementary evaluation metrics tailored for regression problems. 
We use the mean absolute error (MAE $\downarrow$), mean square error (MSE $\downarrow$), and coefficient of determination (R$^2$ $\uparrow$) as our primary evaluation metrics. Here, $\downarrow$ indicates that lower values have better performance, while $\uparrow$ signifies that higher values are preferred.
These metrics are formally defined as follows:

\begin{equation*}
    \begin{aligned}
        \text{MAE}&=\frac{1}{n}\sum_{i=1}^n|\hat{y}_i-y_i|,\\
        \text{MSE}&=\frac{1}{n}\sum_{i=1}^n(\hat{y}_i-y_i)^2,\\
        R^2&=1-\dfrac{\sum_{i=1}^n(\hat{y}_i-y_i)^2}{\sum_{i=1}^n(\hat{y}_i-\bar{y})^2},
    \end{aligned}
\end{equation*}
where $\hat{y}_i$ is the predicted economic indicator value for the $i$-th sample, $y_i$ is the corresponding ground truth value, and $\bar{y}$ is the mean value of all ground truth observations in the dataset.

\subsection{Model performance on GDP mapping task}

\subsubsection{Sensitivity analysis of the loss factor}

In this experiment, we systematically varied the loss factor $\lambda$, which balances self-supervision and direct supervision in the SemiGTX loss function (Eq. \ref{eq:loss}), testing values of \{0, 0.1, 0.3, 0.5, 0.7, 0.9\}.
Self-supervision enables the model to learn spatial relationships between subdistrict-level nodes to obtain spatially interacted representations, while supervision involves local regression to improve the mapping accuracy for individual districts. Higher $\lambda$ values emphasize self-supervision (spatial interaction learning), whereas lower values prioritize direct supervision (local mapping performance). Table \ref{tab:factors} presents the comparative results.

\begin{table*}[!htbp]
    % \vspace{-10pt}
    \caption{Model performance on different loss factors}
    \label{tab:factors}
    \centering
    % \small
    \fontsize{10pt}{10pt}\selectfont
    \resizebox{\linewidth}{!}{
    \begin{tabular}{cccccccccc}\toprule
    \multirow{2}{*}{\textbf{Loss factor}} 
    &\multicolumn{3}{c}{\textbf{Primary}} &\multicolumn{3}{c}{\textbf{Secondary}} &\multicolumn{3}{c}{\textbf{Tertiary}}\\\cmidrule{2-4}\cmidrule{5-7}\cmidrule{8-10}
    &\textbf{MAE $\downarrow$} &\textbf{MSE $\downarrow$} &\textbf{R$^2$ $\uparrow$}
    &\textbf{MAE $\downarrow$} &\textbf{MSE $\downarrow$} &\textbf{R$^2$ $\uparrow$}
    &\textbf{MAE $\downarrow$} &\textbf{MSE $\downarrow$} &\textbf{R$^2$ $\uparrow$}\\\midrule

    \textit{0} & 16.30 & 419.26 & 0.64 & 191.68 & 131942.11 & 0.90 & 338.91 & 232512.67 & 0.89\\
    0.1 & 11.87 & 318.12 & 0.73 & 213.67 & 143384.89 & 0.89 & 282.15 & 135879.42 & 0.93\\
    0.3 & 9.91 & 175.65 & 0.85 & 160.39 & 46329.87 & 0.96 & \textbf{206.30} & 145789.48 & 0.93\\
    0.5 & \textbf{5.98} & \textbf{85.36} & \textbf{0.93} & 190.05 & 54949.89 & 0.96 & 253.08 &\textbf{ 116385.90} & \textbf{0.94}\\
    0.7 & 8.78 & 140.68 & 0.88 & 180.23 & 78247.18 & 0.94 & 403.15 & 492092.34 & 0.76\\
    0.9 & 9.30 & 159.88 & 0.86 & \textbf{149.16} & \textbf{41549.47} & \textbf{0.97} & 203.22 & 146512.46 & 0.93\\
    \bottomrule
    \end{tabular}
    }
    \vspace{-5pt}
    % \vspace{5pt}
\end{table*}

The results reveal that when only supervised learning is considered (setting $\lambda=0$), the model adequately captures the mappings for the secondary and tertiary sectors, achieving R² values of 0.90 and 0.89, respectively. However, it performs considerably worse for the primary sector, with an R² of only 0.64.
Incorporating self-supervision ($\lambda > 0$) significantly enhances performance for the primary sector, with R² increasing from 0.64 to a maximum of 0.93, while performance for the secondary and tertiary sectors remains relatively stable.
Among the values tested, $\lambda=0.5$ provides the best overall performance, achieving R² scores of 0.93, 0.96, and 0.94 for the primary, secondary and tertiary sectors, respectively.

In general, the primary sector, which involves agriculture, forestry, and related industries, proves difficult to model using only local information. Self-supervision, which improves understanding of spatial interactions, effectively addresses this limitation. 
Our experiments demonstrate that combining balanced self-supervision with direct supervision provides the model with complementary perspectives, substantially improving its mapping capabilities.

\subsubsection{The selection of P/S encoder types}

Table \ref{tab:pese} presents a detailed comparison of the evaluation metrics in the three economic sectors when employing different types of P/S encoders, while maintaining all other settings consistent with the full SemiGTX model. All comparative experiments use $\lambda=0.5$.
The results indicate that LapPE significantly improves performance in the secondary sector, achieving an R² of 0.90 compared to 0.86 when using RWSE. Conversely, RWSE exhibits stronger performance in the primary sector, with an R² of 0.89 compared to LapPE's 0.85.
Additionally, implementing both encoders (LapPE and RWSE) simultaneously, compared to omitting all P/S encoders (\textit{none}), significantly improves the model's accuracy across all three sectors, achieving nearly a 20\% performance enhancement.
This underscores the critical importance of incorporating positional and structural encodings. Different encodings benefit specific economic sectors, and their combination yields a more robust model capable of capturing complex spatial and structural relationships within the data.

\begin{table*}[!hbtp]
    % \vspace{-10pt}
    \caption{Ablation study of the P/S encoders}
    \label{tab:pese}
    \centering
    % \small
    \fontsize{10pt}{10pt}\selectfont
    \resizebox{\linewidth}{!}{
    \begin{tabular}{cccccccccc}\toprule
    \multirow{2}{*}{\textbf{P/S encoders}} 
    &\multicolumn{3}{c}{\textbf{Primary}} &\multicolumn{3}{c}{\textbf{Secondary}} &\multicolumn{3}{c}{\textbf{Tertiary}}\\\cmidrule{2-4}\cmidrule{5-7}\cmidrule{8-10}
    &\textbf{MAE $\downarrow$} &\textbf{MSE $\downarrow$} &\textbf{R$^2$ $\uparrow$}
    &\textbf{MAE $\downarrow$} &\textbf{MSE $\downarrow$} &\textbf{R$^2$ $\uparrow$}
    &\textbf{MAE $\downarrow$} &\textbf{MSE $\downarrow$} &\textbf{R$^2$ $\uparrow$}\\\midrule
    \textit{none} & 8.49 & 271.09 & 0.77 & 326.53 & 268643.77 & 0.79 & 406.16 & 505506.57 & 0.76\\
    LapPE & 12.26 & 227.84 & 0.81 & 195.3 & 121057.05 & 0.90 & 374.56 & 319537.03 & 0.85\\
    RWSE & 8.89 & 179.78 & 0.85 & 165.34 & 179298.52 & 0.86 & 338.91 & 232512.67 & 0.89\\
    LapPE+RWSE & 5.98 & 85.36 & 0.93 & 190.05 & 54949.89 & 0.96 & 253.08 & 116385.90 & 0.94\\
    \bottomrule
    \end{tabular}
    }
    \vspace{-5pt}
    % \vspace{5pt}
\end{table*}

\subsubsection{The roles of input modalities}

A distinctive feature of SemiGTX is its ability to process multi-modal input data. To rigorously assess the impact of different data modalities on model performance, we set $\lambda=1$, which allows the model to learn feature representations entirely through self-supervision. We then added several standard regressors—Ridge Regression, Support Vector Machines (SVM), Random Forest (RF) and Multilayer Perceptron (MLP)—to evaluate the contributions of SVIs and POIs. This approach eliminates potential biases in the projection that might arise from the supervision component in SemiGTX.

We organized the input features into three scenarios based on the data modality: POI-only, SVI-only, and both modalities combined. All other parameters remained consistent with the full SemiGTX model. The use of multiple regressors helps mitigate the bias introduced by relying on a single regressor.
As shown in Table \ref{tab:regressors}, considering the average performance across all regressors, POI significantly enhance the mapping accuracy for the secondary and tertiary sectors, while SVI are more advantageous to map the primary sector. Using the complementary strengths of both modalities, we achieve superior overall performance.

\begin{table*}[!hbtp]
    % \vspace{-10pt}
    \caption{Evaluation of POIs, SVIs, and different regressors}
    \label{tab:regressors}
    \centering
    \resizebox{\linewidth}{!}{
    % \small
    \fontsize{10pt}{10pt}\selectfont
    \begin{tabular}{ccccccccccc}\toprule
    \multirow{2}{*}{\textbf{Features}} & \multirow{2}{*}{\textbf{Regressors}} 
    &\multicolumn{3}{c}{\textbf{Primary}} &\multicolumn{3}{c}{\textbf{Secondary}} &\multicolumn{3}{c}{\textbf{Tertiary}}\\\cmidrule{3-5}\cmidrule{6-8}\cmidrule{9-11}
    &&\textbf{MAE $\downarrow$} &\textbf{MSE $\downarrow$} &\textbf{R$^2$ $\uparrow$}
    &\textbf{MAE $\downarrow$} &\textbf{MSE $\downarrow$} &\textbf{R$^2$ $\uparrow$}
    &\textbf{MAE $\downarrow$} &\textbf{MSE $\downarrow$} &\textbf{R$^2$ $\uparrow$}\\\midrule
    \multirow{4}{*}{\text{POI-only}}
    &Ridge &16.06 &540.13 &0.54 &289.92 &208483.86 &0.84 &506.81 &729681.36 &0.65\\
    &SVM   &16.98 &603.07 &0.47  &348.21 &368547.52 &0.71   &586.90 &1336308.36 &0.36 \\
    &RF    &12.82 &312.34 &0.72  &257.90 &249926.14 &0.80  &379.66 &491284.72 &0.76 \\
    &MLP &20.23 &680.46 &0.42 &347.69 &245834.09 &0.81 &604.81 &1139688.22 &0.45\\\midrule
    \multirow{4}{*}{\text{SVI-only}}
    &Ridge &9.89 &175.99 &0.85 &636.86 &796567.51 &0.45 &325.01 &652348.76 &0.71\\
    &SVM   &9.59 &151.92 &0.86 &404.73 &505582.70 &0.61 &458.92 &887316.02 &0.57 \\
    &RF    &23.65 &799.99 &0.29 &702.55 &965857.80 &0.24   &745.38 &1250170.60 &0.40 \\
    &MLP &9.03 &167.14 &0.86 &731.68 &1027657.66 &0.26 &424.40 &682745.41 &0.66\\\midrule
    \multirow{4}{*}{\text{Both}}
    &Ridge &16.92 &467.07 &0.60 &388.81 &372345.99 &0.71 &352.83 &336764.19 &0.84\\
    &SVM   &18.80 &780.34 & 0.32 &307.12 &281417.70 & 0.77  &524.71 &1193403.50 & 0.43\\
    &RF  &11.76 &276.73 & 0.75 &275.48  &316929.40 & 0.75  &418.48 &587772.40 & 0.72\\
    &MLP & 18.75 &520.30 &0.55 &422.30 &372395.89 &0.70 &426.18 &400370.98 &0.81\\
    \bottomrule
    \end{tabular}
    }
    \vspace{-5pt}
    % \vspace{5pt}
\end{table*}

\subsubsection{Comparison with baseline models}

We compared SemiGTX with several baseline methods, including classical Graph Neural Networks (GNNs) like GCN, GatedGCN, and GAT; Transformer-based networks such as Transformer, BigBird, and Performer; contrastive learning methods such as SimCLR and GIM; and SemiGTX variants.
Specifically, SemiGTX-GT replaces the MPNN layer in SemiGTX with a GCN and the attention layer with a Transformer, while SemiGTX-ViT uses a pre-trained ViT as the image pre-encoder for SVIs instead of our fine-tuned ViT. All these SemiGTX variants maintain $\lambda=0.5$.

\begin{table*}[!htbp]
    % \vspace{-10pt}
    \caption{Comparison with baseline methods on GDP mapping task}
    \label{tab:layers}
    \centering
    \resizebox{\linewidth}{!}{
    % \small
    \fontsize{10pt}{10pt}\selectfont
    \begin{tabular}{cccccccccc}\toprule
    \multirow{2}{*}{\textbf{Model}} 
    &\multicolumn{3}{c}{\textbf{Primary}} &\multicolumn{3}{c}{\textbf{Secondary}} &\multicolumn{3}{c}{\textbf{Tertiary}}\\\cmidrule{2-4}\cmidrule{5-7}\cmidrule{8-10}
    &\textbf{MAE $\downarrow$} &\textbf{MSE $\downarrow$} &\textbf{R$^2$ $\uparrow$}
    &\textbf{MAE $\downarrow$} &\textbf{MSE $\downarrow$} &\textbf{R$^2$ $\uparrow$}
    &\textbf{MAE $\downarrow$} &\textbf{MSE $\downarrow$} &\textbf{R$^2$ $\uparrow$}\\\midrule
    GCN & 16.75 & 568.58 & 0.52 & 553.77 & 683070.43 & 0.46 & 989.91 & 1376486.22 & 0.33\\
    GatedGCN & 13.99 & 413.44 & 0.63 & 439.06 & 424493.89 & 0.67 & 1073.76 & 1614047.39 & 0.22\\
    GAT & 16.76 & 445.47 & 0.62 & 428.04 & 428897.01 & 0.66 & 567.92 & 837465.95 & 0.60\\\midrule
    Transformer & 19.05 & 893.31 & 0.24 & 372.75 & 282106.92 & 0.78 & 368.18 & 455634.82 & 0.78\\
    BigBird & 23.74 & 1003.79 & 0.14 & 274.11 & 235878.06 & 0.81 & 344.39 & 412747.10 & 0.80\\
    Performer & 19.95 & 791.42 & 0.33 & 165.84 & 179308.11 & 0.86 & 310.05 & 336238.73 & 0.84\\\midrule
    SimCLR & 9.30 & 159.88 & 0.86 & 242.29 & 283787.94 & 0.78 & 530.43 & 513237.08 & 0.75\\
    GIM  & 9.87 & 165.63 & 0.86 & 325.90 & 268775.66 & 0.79 & 343.43 & 330628.81 & 0.84\\
    SemiGTX-GT & 8.36 & 134.96 & 0.88 & 179.03 & 136391.49 & 0.89 & 308.73 & 232635.82 & 0.89\\\midrule
    SemiGTX-ViT & 8.15 & 126.3 & 0.89 & 170.93 & 81819.94 & 0.94 & 288.36 & 209558.51 & 0.90\\
    SemiGTX & 5.98 & 85.36 & 0.93 & 190.05 & 54949.89 & 0.96 & 253.08 & 116385.90 & 0.94\\
    \bottomrule
    \end{tabular}
    }
    \vspace{-5pt}
    % \vspace{5pt}
\end{table*}

As shown in Table \ref{tab:layers}, all graph neural networks demonstrate stronger performance in the primary and secondary sectors compared to the tertiary sector, indicating their effectiveness in learning spatial interaction patterns. This makes them particularly suitable for industries like agriculture and manufacturing, which cannot be adequately characterized by local information alone. Among these, GAT achieves the best results, with R² values exceeding 0.60 in the three economic sectors. In contrast, transformer-based networks perform poorly for the primary sector, with R² values below 0.40 for the three Transformer variants.
The contrastive learning models (SimCLR and GIM), which incorporate GCN and Transformer architectures similar to SemiGTX-GT, show limited sectoral performance due to their heavy emphasis on self-supervision without adequate supervisory signals. By comparison, SemiGTX-GT, which integrates the supervision benefits, shows improved overall performance.
The SemiGTX-ViT variant, which employs a non-fine-tuned image pre-encoder, performs worse than the original SemiGTX across all economic sectors that heavily rely on visual semantic information.
Overall, the fine-tuned image pre-encoder in SemiGTX provides more robust semantic representations of street view images while combining the strengths of multiple modeling approaches, making it the superior choice among all baseline methods.

\subsubsection{The cross-regional generality}

To rigorously assess the generalization capability of SemiGTX, we conducted validation studies in two cities with contrasting economic and geographic profiles: Beijing and Chengdu. 
Beijing, as China's capital and service-driven megacity with rich multi-modal data (including high-resolution street view imagery), represents a highly developed urban economy. In contrast, Chengdu, located in western China, exemplifies a transitional city with diverse economic sectors and variable data quality, reflecting the complexities of emerging urban regions. \ref{app2} Table \ref{tab:bjcd} details the disparities in multi-source data volumes between Beijing and Chengdu. Compared to Beijing, Chengdu exhibits lower data availability across multi-source modalities, particularly in high-resolution SVIs and O-D records.

\begin{table*}[!htbp]
    \caption{Performance comparison across additional regions}
    \label{tab:cross}
    \centering
    \resizebox{\linewidth}{!}{
    % \small
    \fontsize{10pt}{10pt}\selectfont
    \begin{tabular}{lcccccccccc}\toprule
    \multirow{2}{*}{\textbf{Regions}} & \multirow{2}{*}{\textbf{Models}} 
    &\multicolumn{3}{c}{\textbf{Primary}} &\multicolumn{3}{c}{\textbf{Secondary}} &\multicolumn{3}{c}{\textbf{Tertiary}}\\\cmidrule{3-5}\cmidrule{6-8}\cmidrule{9-11}
    &&\textbf{MAE $\downarrow$} &\textbf{MSE $\downarrow$} &\textbf{R$^2$ $\uparrow$}
    &\textbf{MAE $\downarrow$} &\textbf{MSE $\downarrow$} &\textbf{R$^2$ $\uparrow$}
    &\textbf{MAE $\downarrow$} &\textbf{MSE $\downarrow$} &\textbf{R$^2$ $\uparrow$}\\\midrule
    \multirow{3}{*}{Beijing} 
    & GAT & 2.13 & 10.85 & 0.69 & 72.31 & 11830.16 & 0.76 & 1119.46 & 3115425.75 & 0.64\\
    & SimCLR & 1.98 & 7.58 & 0.78 & 47.38 & 6018.46 & 0.88 & 674.42 & 1204975.5 & 0.86\\
    & SemiGTX & 1.77 & 6.24 & 0.82 & 55.58 & 5234.98 & 0.89 & 576.04 & 622656.79 & 0.93\\\midrule
    \multirow{3}{*}{Chengdu} 
    & GAT & 13.99 & 413.44 & 0.63 & 97.03 & 21735.37 & 0.50 & 258.09 & 107820.25 & 0.81\\
    & SimCLR & 7.87 & 143.06 & 0.72 & 87.25 & 15431.84 & 0.65 & 153.29 & 50129.45 & 0.91\\
    & SemiGTX & 7.25 & 106.95 & 0.79 & 74.82 & 8664.94 & 0.80 & 141.21 & 39411.51 & 0.93\\\midrule
    \bottomrule
    \end{tabular}
    }
    \vspace{-5pt}
\end{table*}

As demonstrated in Table~\ref{tab:cross}, SemiGTX achieves impressive R² scores of 0.82 (primary), 0.89 (secondary) and 0.93 (tertiary) in Beijing. These strong results are driven by its effective feature extraction capabilities across multiple data modalities, as detailed in Section 4.3.3. In Chengdu, despite more challenging data conditions characterized by data sparsity, the model achieves robust performance with R² scores of 0.79, 0.80, and 0.93 in the three sectors. This resilience can be attributed to the model's semi-supervised loss mechanism, which effectively leverages limited labeled data while capturing underlying spatial patterns. Furthermore, experiments in both cities consistently demonstrate that SemiGTX outperforms baseline methods (with GAT and SimCLR configured as described in Section 4.3.4). Collectively, these cross-regional validation results confirm SemiGTX's strong generalization capabilities, demonstrating its effectiveness in data-rich environments while maintaining robust performance under more challenging data-sparse conditions.
This versatility makes SemiGTX particularly valuable for economic mapping in diverse urban contexts with variable data availability and quality.

\section{Discussion}\label{s5}

In this section, we provide an in-depth analysis of the model's performance, with particular emphasis on spatial heterogeneity effects. We also examine how different input data modalities contribute to the mappings across the three economic sectors, thereby enhancing the model's explainability.

\subsection{Spatial heterogeneity effects}

\begin{figure}[t]
	\centering
		\includegraphics[width=\textwidth]{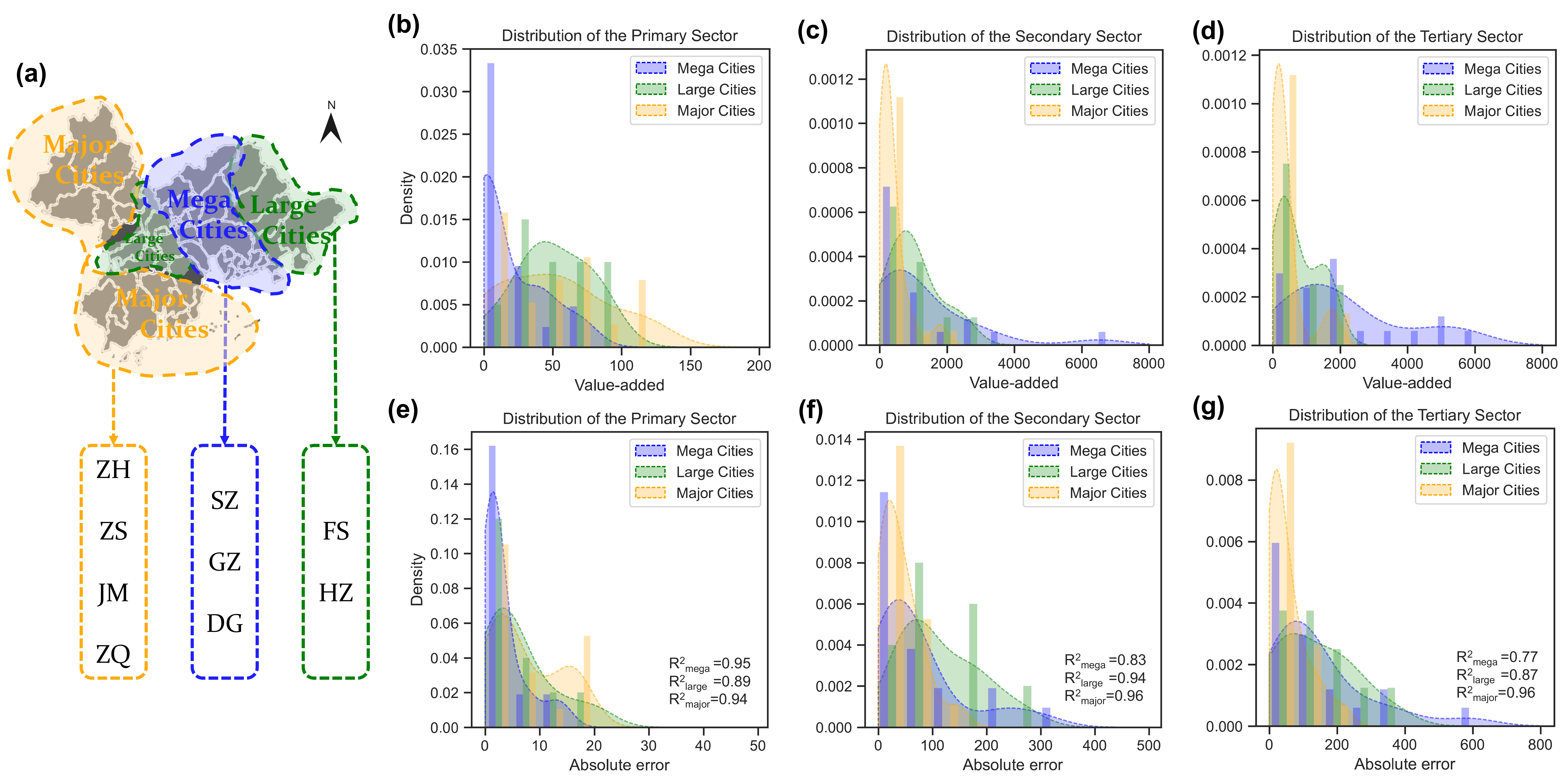}
    \caption{Distribution of the three economic sectors in the mega cities, large cities, and major cities. (a) Schematic division of the PRD Region into three groups. (b-d) Density distribution of value-added in the three economic sectors. (e-g) Density distribution of model absolute error in the three economic sectors.}\label{fig:citydist}
\end{figure}

To visually illustrate spatial heterogeneity, we adopted the city-scale classification standards provided by the State Council of China to categorize cities in the PRD into three groups based on population: \textbf{mega cities} (populations over 10 million, including Shenzhen (SZ), Guangzhou (GZ), and Dongguan (DG)), \textbf{large cities} (populations between 5 to 10 million, including Foshan (FS) and Huizhou (HZ)), and \textbf{major cities} (populations between 1 to 5 million, including Zhuhai (ZH), Zhongshan (ZS), Jiangmen (JM), and Zhaoqing (ZQ)).
Figure \ref{fig:citydist}(a) shows the spatial distribution of these three groups.

% Figure

% Figure
\begin{figure}[t]
	\centering
		\includegraphics[width=\textwidth]{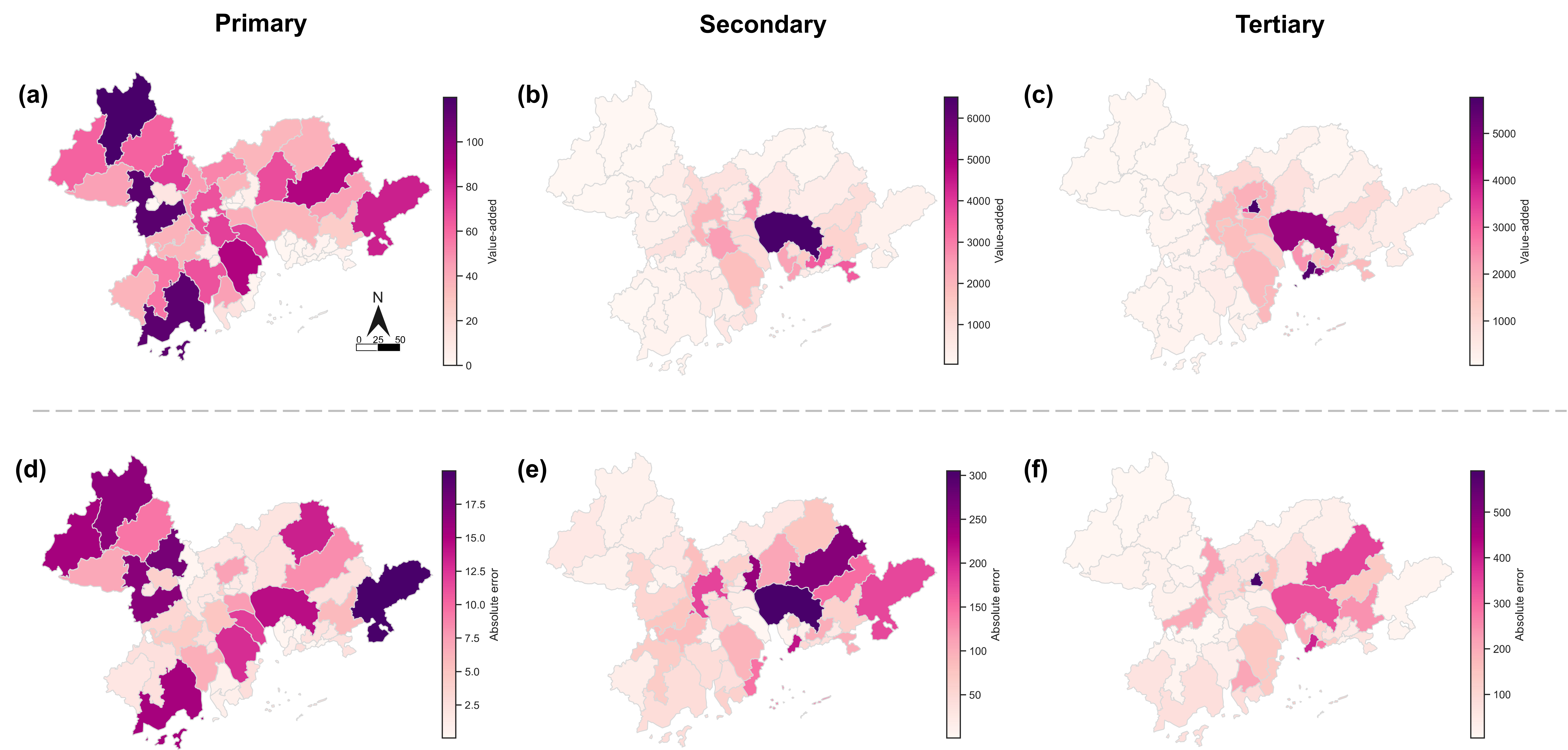}
    \caption{Geographic distribution of (a-c) value-added and (d-f) model absolute error in the three economic sectors.}\label{fig:density}
\end{figure}

Figures \ref{fig:citydist}(b-d) and Figures \ref{fig:density}(a-c) illustrate the density and geographic distributions of value-added across the primary, secondary and tertiary sectors between different city groups. 
In the primary sector, value-added is generally low in \textbf{mega cities}, whereas some \textbf{major cities} exhibit higher value-added, with \textbf{ large cities} positioned intermediately.
Conversely, in the secondary and tertiary sectors, \textbf{major cities} typically display the lowest value-added, \textbf{large cities} have higher value-added than \textbf{major cities}, and \textbf{mega cities} cover a wider range, including the highest value-added figures within these economic sectors.
The geographical distribution maps also clearly show that higher value-added in the secondary and tertiary sectors is concentrated in the central \textbf{mega cities}, diminishing progressively toward peripheral areas.
These observations highlight significant differences in industrial structure among the city groups and highlight the tendency of populations, which closely correlate with GDP, to concentrate in cities with more developed secondary and tertiary sectors.

Figures \ref{fig:citydist}(e-g) and Figures \ref{fig:density}(d-f) show the density and spatial distributions of absolute error between our SemiGTX model predictions and ground truth values across the three economic sectors. Direct comparison between value-added distributions and their corresponding error distributions (Figures \ref{fig:density}(a) and (d), (b) and (e), (c), and (f)) reveals clear similarities.
Specifically, the model performs better for groups with high density and lower values, thus exhibiting a reduced mapping performance for districts in cities with exceptionally high values. For example, R² values for the primary sector in the \textbf{large cities} group are lower than in other groups, while the R² values for the secondary and tertiary economic sectors are also lower in the \textbf{mega cities} group compared to other groups. 

This distinct pattern stems from inherent spatial heterogeneity, as evidenced by the consistently observed long-tailed characteristics in the density distributions.
The observed heterogeneous GDP development patterns arise from differences in historical evolution, resource availability, and environmental conditions between districts. These disparities shape the economic growth trajectories and GDP outcomes of each district within the groups, leading to significant variations in GDP levels. 
Consequently, spatial evolutions and industrial structures differ markedly between the different city groups\citep{wei2023spatial}.
This characteristic mirrors the behavior of the ground truth data and highlights that our model effectively captures this fundamental spatial heterogeneity.

\subsection{Multi-modal explainability}\label{s5s1}

% Figure
\begin{figure}[t]
	\centering
		\includegraphics[width=\textwidth]{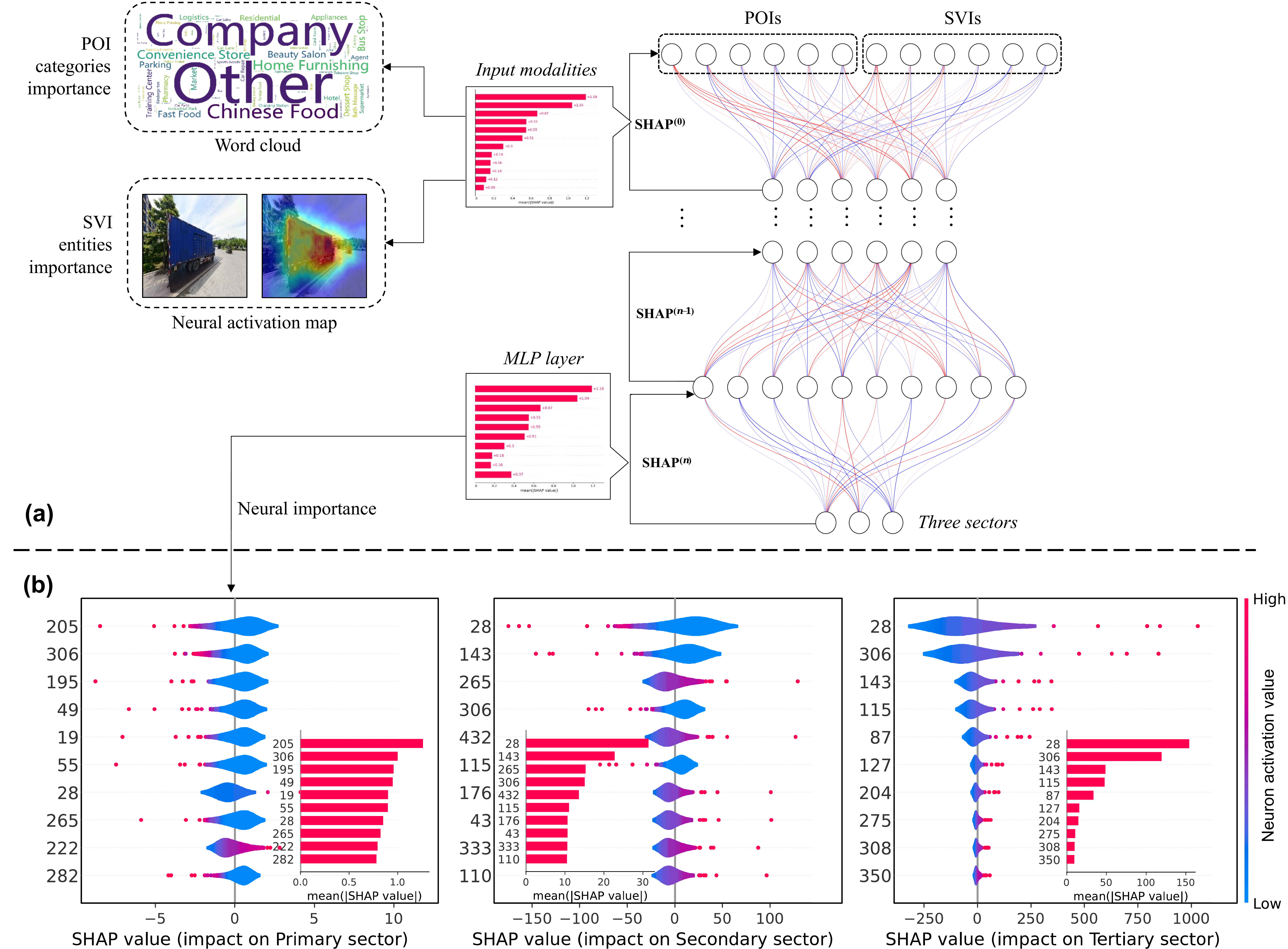}
	  \caption{SHAP values on the GDP mapping task. (a) Schematic of layer-wise trace-backing for SHAP calculation, (b) SHAP values of the top 10 neurons for the three sectors. Insert: Overall importance of neurons, represented by the mean of absolute SHAP values}\label{fig:traceback}
\end{figure}

By applying the SHAP (SHapley Additive exPlanations) method \citep{lundberg2017unified}, a game-theoretic approach that attributes model predictions to input features, we traced back through each layer of our model to determine how different input modalities contribute to the final outputs.
Given SemiGTX's complex network structure, directly determining the importance of input features from the output layer alone is not feasible. 
To address this challenge, we employ a layer-wise tracing strategy to calculate SHAP values, as illustrated in Figure \ref{fig:traceback}(a).
Starting from the last layer (the $n$-th layer) of the neural network, we first compute the SHAP values for each neuron in the MLP layer to identify the importance of the top $k$ neurons with the highest impact. Using these top neurons as references, we then trace back to the preceding layer (the ($n$-1)-th layer). This iterative process continues through each layer until we reach the input layer, ultimately allowing us to determine the significance of different input modalities contributing to the model predictions.

Figure \ref{fig:traceback}(b) illustrates the influence of the top 10 neurons in the MLP layer that have the highest impact on the final mapping results in all samples for the three economic sectors. The beeswarm plots display SHAP values along the x-axis, indicating each neuron's contribution: A SHAP value $>0$ indicates a positive effect on the output, while a SHAP value $<0$ indicates a negative effect. The inset bar plots show the means of the absolute SHAP values, representing the overall importance of each neuron. The y-axis of all plots is sorted by neuron importance. It is evident that different neurons predominantly control the mapping results for different economic sectors. Specifically, for the primary sector, most neurons have a positive effect (SHAP value $>0$) when their activation values are lower. Most samples correspond to these neurons with low activation values, which is reflected by the larger spread of points at lower activation levels in the plot. In contrast, for the tertiary sector, most neurons have a positive effect when their activation values are higher. Notably, neurons 28, 143, and 306 exhibit mutually exclusive effects on the secondary and tertiary economic sectors: higher activation values in these neurons suppress the secondary sector(negative SHAP values) but enhance the tertiary sector(positive SHAP values). By analyzing these calculated SHAP values, our approach provides a clear and interpretable explanation of the model's behavior, facilitating the validation of its reliability and the identification of potential biases.

\begin{figure}[t]
	\centering
    \includegraphics[width=\textwidth]{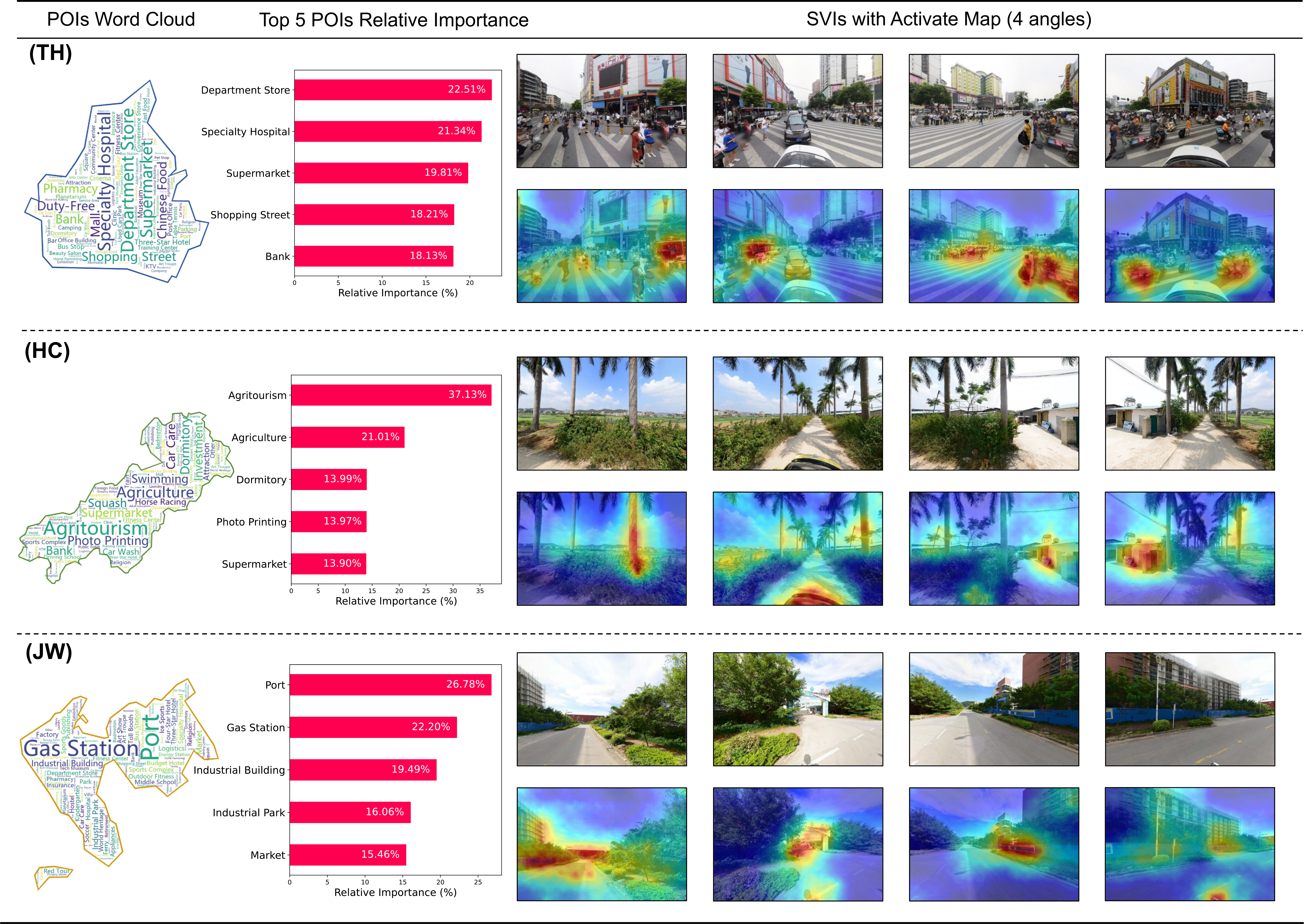}
    \caption{Importance of input modalities across the three city groups, visualized using POI word clouds, Top 5 POIs relative importance, and SVIs with activation maps (4 Views).  (TH) A case in Tianhe District, Guangzhou. (HC) A case in Huicheng District, Huizhou. (JW) A case in Jinwan District, Zhuhai.}
   \label{fig:input}
\end{figure}

To enhance our understanding of how various input modalities influence outcomes, we utilize word clouds for POIs to emphasize the key categories within the POI dataset. For illustrating the significance of SVIs, we adapt the Class Activation Mapping (CAM) method \citep{selvaraju2017grad} to detect and accentuate visual elements on which the dominant neuron of the second layer - specifically the neuron with the highest SHAP value—concentrates on. We term this approach the Neural Activation Map (NAM).
NAM generates heat maps that depict the responses of the back-propagation gradient of hidden layers to specific neurons. By backpropagating the predicted outputs through the network, we obtain gradient information for the specified hidden layer. The regions with higher response values, indicated by red pixel clusters in the heat map, represent the entities the model considers most important, as shown in Figure \ref{fig:traceback}(a).

Starting with neurons 205 and 28—the top neurons in the final layer—we traced back through each layer to the input layer, computing SHAP values at each step. By averaging these values, we determined the average importance of different input modalities (POIs and SVIs) for these neurons in the sample districts of different city groups. Figure \ref{fig:input} illustrates the importance of input modalities for three selected districts: Tianhe (TH) in Guangzhou, Huicheng (HC) in Huizhou and Jinwan (JW) in Zhuhai, each representing one of the three city groupings.
The POI word cloud \footnote{The boundary of the word cloud corresponds to the city's borders, but for clearer visualization, some very small islands are omitted.} displays the importance of different POI categories, with larger font sizes indicating higher importance. 
The bar plots present the relative importance of the five most important POI categories. 
Finally, sample SVIs captured from four different angles within the district, along with their corresponding activation maps, highlight the visual entities that have the greatest impact on the output.

Figure \ref{fig:input} (TH) presents the results for Tianhe, Guangzhou, a representative district in the \textbf{mega cities} group. POI categories related to the tertiary sector, such as \textit{Department Store} and \textit{Specialty Hospital}, show higher importance. 
The SVI sample depicts a bustling street scene near a market, where our model accurately captures dense pedestrian traffic and commercial characteristics. These findings align with our previous analysis that \textbf{mega cities} typically exhibit higher value-added in the tertiary sector.
Figure \ref{fig:input} (JW) shows Jinwan in Zhuhai, exemplifying the \textbf{major cities} group. Among the POI categories, those related to the secondary sector, such as \textit{Port} and \textit{Gas Station}, are particularly prominent. The SVI sample emphasizes entities such as factories and construction sites, which align well with the prominent POI categories. This correspondence underscores the significant role of secondary sector activities in major cities, reflecting their focus on industrial production and manufacturing.
Figure \ref{fig:input} (HC) shows the results for Huicheng in Huizhou, representing the \textbf{large cities} group. Here, the most important POI categories are associated with the primary sector, like \textit{Agritourism} and \textit{Agriculture}. The SVI sample also highlights trees, fields, and small houses, aptly reflecting the landscape characteristics of \textbf{large cities}. These elements indicate a mixture of agricultural activity and burgeoning industrial development typical of the cities in this group.

Overall, our analysis reveals how different data modalities influence SemiGTX's predictions across the three economic sectors, providing insight into the model's decision-making process. 
Figure~\ref{fig:input} illustrates key regional features and their processing through the framework.
By examining industrial structures and analyzing the impact of POI categories and SVI elements across districts, we identify meaningful relationships that explain how our model differentiates between economic sectors, enhancing its explainability.

\subsection{Limitations and future work}

SemiGTX proposes a semi-supervised graph representation learning to improve the spatial representation of multimodal data and mapping of various economic indicators. Despite these contributions, several limitations and future directions remain.

First, the current model can be further improved by exploring more advanced embedding and encoding techniques, as well as incorporating other intelligent methods for multi-task learning and semi-supervised learning. 
For example, recent advances in POI embedding techniques that incorporate temporal dynamics \citep{yao2023unsupervised} can capture complex and dynamic relationships and patterns in POI data. 
Furthermore, emerging neural network architectures and multi-modal fusion methods \citep{liu2024capsule, liu2025part} also hold great promise for enhancing our SemiGTX framework. The integration of other advanced graph-based semi-supervised learning methods \citep{qiao2024information, sun2024graph}, distinct from the approach used in this study, is also worth exploring.

Second, the cost-effective transferability of SemiGTX across various regions remains underexplored. Closing this gap is essential for enhancing the model's broad applicability. Regions differ in spatial heterogeneity and possess unique socio-economic traits that can influence model performance (as illustrated by the comparison between Beijing and Chengdu in Section 5.3.5). Future research could focus on assessing the model's domain adaptation capabilities in diverse urban settings, such as the Yangtze River Delta, Northeast China, and international locations with distinct regional profiles. Leveraging techniques like transfer learning and domain adaptation \citep{shi2024graph, qiao2024information} could be particularly beneficial, enabling the model to apply knowledge gained from one region to improve its effectiveness in others.

\section{Conclusions}\label{s6}
The SemiGTX framework advances multi-modal geospatial big data integration, specifically designed to enhance economic indicator mapping in semi-supervised learning contexts by strategically balancing self-supervision and direct supervision.
SemiGTX employs modality-specific encoding architectures for each input type, integrating representations in a weighted graph structure. GraphGPS synthesizes these features into a holistic representation. Leveraging multi-task supervision with limited labeled data while optimizing subgraph-level mutual information enables robust spatial dependency modeling despite data sparsity. This innovation seamlessly integrates labeled and unlabeled data streams to optimize performance, addressing the fundamental challenge of limited labeled data in economic forecasting while demonstrating enhanced robustness and superior generalization across diverse applications.

Experiments demonstrate SemiGTX's effectiveness, achieving remarkable R² values of 0.93, 0.96, and 0.94 for GDP distribution across primary, secondary, and tertiary economic sectors in the Pearl River Delta region.
Ablation studies further confirm that strategically balancing spatial interaction relationships with local supervision signals significantly boosts model performance by an average of 17.2\%. The combined use of SVIs and POIs provides richer multidimensional information compared to using them separately, resulting in a 6.8\% improvement in prediction accuracy. 
Additionally, cross-regional validation in Beijing and Chengdu achieves average R² values of 0.88 and 0.84, respectively, further confirming SemiGTX's SemiGTX's robust generalization capabilities and its effectiveness under limited data conditions.

Through SHAP attribution analysis, we quantify the impact of multi-source input features on model outputs, enhancing the SemiGTX's explainability and providing valuable insights for regional development planning. We also address potential validity concerns, including data quality, availability, and learning methodologies. Future work will focus on designing more advanced semi-supervised learning strategies and developing large models that adhere to transfer learning paradigm to address cross-regional performance challenges.

\newpage
\appendix

\setcounter{table}{0}

\section{ViT Pre-encoder Fine-tuning Details}\label{app1}

\begin{table}[!htbp]
    \centering
    \caption{The hyperparameter configurations of ViT encoder}
    \fontsize{10pt}{10pt}\selectfont
    \begin{tabular}{ccc}
    \toprule[1.0pt] %\hline
        \multirow{1}{*}{} & Parameter & Value \\ \toprule[1.0pt] %\hline
        \multirow{4}{*}{Global}  & Patch size & \(16\times 16\)\\
         &Latent dimension & 768\\
         &MLP dimension & \(768\times 4\)\\
         &Channels & 3\\ \hline
         \multirow{2}{*}{Decoder} &Attention heads & 6\\
         &Transformer depth & 4\\
    \toprule[1.0pt] %\hline
    \end{tabular}
    \label{tab:config}
\end{table}

The detailed hyperparameter configurations for the additional ViT decoder are outlined in Table \ref{tab:config}. 
To optimize training efficiency, street view images are first rescaled to $512\times 512$ resolution before being fed into the encoder. The initial learning rate is 0.0001, with a decaying rate of 0.5 after every 100 epochs. The training encompasses 1000 epochs, with each epoch consisting of 16 batches, executed in a Linux environment leveraging a single NVIDIA A6000 GPU.

\newpage
\section{Statistical Summary of Raw Data}\label{app2}

\setcounter{table}{0}
\setcounter{figure}{0}

{\small\tabcolsep=3pt  % hold it local
\fontsize{9pt}{9pt}\selectfont
\begin{longtable}{c|ccccc}

    \caption{GDP of Primary, Secondary, and Tertiary economic sectors of Districts in the Pearl River Delta.} \label{tab:gdp} \\\hline 
    
    \multirow{2}{*}{\textbf{District}} & \multicolumn{3}{c}{\textbf{GDP} (100 million CNY)} & \multirow{2}{*}{\textbf{Area} (km\(^2\))} & \multirow{2}{*}{\textbf{Population} (10000 per)}\\
    & Primary & Secondary & Tertiary \\
    \midrule
    \endfirsthead

    \multicolumn{6}{c}%
    {{\tablename\ \thetable{} -- continued from previous page}} \\\hline 
    \multirow{2}{*}{\textbf{District}} & \multicolumn{3}{c}{\textbf{GDP} (100 million CNY)} & \multirow{2}{*}{\textbf{Area} (km\(^2\))} & \multirow{2}{*}{\textbf{Population} (10000 per)}\\
    & Primary & Secondary & Tertiary \\
    \midrule
    \endhead

    \hline \multicolumn{6}{r}{{Continued on next page}} \\ \hline
    \endfoot
    \hline \hline
    \endlastfoot

    Liwan     & 5.54   & 351.46  & 858.57  & 59.10   & 112.37  \\
    Yuexiu    & 0.00   & 126.05  & 3524.13 & 33.80   & 102.85  \\
    Haizhu    & 1.16   & 450.69  & 2050.68 & 90.40   & 179.83  \\
    Tianhe    & 2.58   & 447.33  & 5765.81 & 96.33   & 222.17  \\
    Baiyun    & 36.55  & 563.98  & 1875.67 & 795.79  & 363.70  \\
    Huangpu   & 5.19   & 2529.15 & 1779.42 & 484.17  & 119.18  \\
    Panyu     & 39.75  & 1016.19 & 1649.53 & 529.94  & 280.74  \\
    Huadu     & 52.06  & 762.20  & 956.54  & 970.04  & 170.62  \\
    Nansha    & 72.28  & 995.50  & 1184.81 & 783.86  & 92.94   \\
    Conghua   & 35.79  & 130.04  & 245.09  & 1974.50 & 73.97   \\
    Zengchen  & 67.40  & 536.71  & 721.16  & 1616.47 & 155.04  \\
    Luohu     & 0.47   & 155.56  & 2474.16 & 78.75   & 101.80  \\
    Futian    & 1.73   & 504.29  & 5008.47 & 78.66   & 151.47  \\
    Nanshan   & 1.08   & 2459.85 & 5574.95 & 187.53  & 181.00  \\
    Baoan     & 0.90   & 2371.83 & 2328.88 & 396.61  & 454.53  \\
    Longgang  & 2.61   & 3460.03 & 1679.57 & 388.22  & 407.36  \\
    Yantian   & 0.32   & 151.43  & 668.87  & 74.99   & 21.15   \\
    Longhua   & 0.60   & 1494.22 & 1456.85 & 175.58  & 249.00  \\
    Pingshan  & 1.25   & 754.27  & 324.12  & 166.31  & 60.87   \\
    Guangming & 2.33   & 1016.97 & 407.81  & 155.44  & 115.09  \\
    Xiangzhou & 2.12   & 938.93  & 1741.17 & 539.76  & 140.66  \\
    Doumen    & 44.62  & 252.65  & 204.00  & 614.32  & 61.71   \\
    Jinwan    & 13.78  & 616.50  & 231.68  & 570.99  & 45.35   \\
    Chanchen  & 0.64   & 792.59  & 1490.58 & 153.88  & 133.07  \\
    Nanhai    & 66.24  & 2006.09 & 1658.26 & 1071.82 & 365.50  \\
    Shunde    & 71.27  & 2478.63 & 1616.49 & 806.57  & 321.44  \\
    Shanshui  & 45.66  & 1068.65 & 358.12  & 827.71  & 87.80   \\
    Gaoming   & 37.32  & 783.85  & 224.01  & 937.81  & 47.42   \\
    Pengjiang & 7.40   & 330.69  & 532.84  & 322.18  & 87.04   \\
    Jianghai  & 7.91   & 184.29  & 109.90  & 109.18  & 37.95   \\
    Xinhui    & 66.31  & 484.67  & 400.65  & 1362.06 & 90.93   \\
    Taishan   & 113.35 & 203.86  & 199.29  & 3308.25 & 89.37   \\
    Kaiping   & 56.50  & 217.57  & 181.99  & 1656.94 & 74.88   \\
    Heshan    & 36.46  & 240.53  & 181.52  & 1082.67 & 54.28   \\
    Enping    & 36.68  & 62.03   & 118.97  & 1693.92 & 48.39   \\
    Duanzhou  & 0.26   & 135.83  & 332.83  & 154.00  & 61.02   \\
    Dinghu    & 11.92  & 70.43   & 70.94   & 552.00  & 21.72   \\
    Gaoyao    & 114.51 & 289.21  & 138.40  & 2185.60 & 74.40   \\
    Guangning & 61.35  & 54.36   & 64.87   & 2455.00 & 40.81   \\
    Huaiji    & 119.81 & 68.52   & 106.35  & 3554.07 & 79.75   \\
    Fengkai   & 61.61  & 58.69   & 50.68   & 2723.93 & 38.01   \\
    Deqing    & 44.30  & 43.68   & 62.96   & 2002.80 & 33.14   \\
    Sihui     & 72.71  & 406.22  & 264.62  & 1267.97 & 65.35   \\
    Huicheng  & 44.86  & 969.30  & 921.48  & 1157.20 & 156.70  \\
    Huiyang   & 25.37  & 1242.13 & 451.73  & 917.13  & 96.81   \\
    Boluo     & 88.50  & 436.48  & 276.41  & 2855.10 & 121.15  \\
    Huidong   & 80.74  & 288.35  & 372.68  & 3528.77 & 102.28  \\
    Longmen   & 37.98  & 83.61   & 81.61   & 2267.20 & 31.94   \\
    Dongguan  & 36.50  & 6513.64 & 4650.18 & 2460.08 & 1048.53 \\
    Zhongshan & 89.20  & 1795.25 & 1746.83 & 1783.67 & 445.82 

\end{longtable}
}

\begin{landscape}
{\small\tabcolsep=3pt  % hold it local
\fontsize{9pt}{9pt}\selectfont
\begin{longtable}{c|cc|cc|c}

    \caption{40 Original POI Data Samples in the Pearl River Delta} \label{tab:op} \\\hline

    \textbf{Name} & \textbf{Category} & \textbf{Subcategory} & \textbf{Lon.} & \textbf{Lat.} & \textbf{District}\\
    \midrule
    \endfirsthead

    \multicolumn{6}{c}%
    {{\tablename\ \thetable{} -- continued from previous page}} \\\hline 
    \textbf{Name} & \textbf{Category} & \textbf{Subcategory} & \textbf{Lon.} & \textbf{Lat.} & \textbf{District}\\
    \midrule
    \endhead

    \hline \multicolumn{6}{r}{{Continued on next page}} \\ \hline
    \endfoot
    \hline \hline
    \endlastfoot
Logistics Phase III Station & Transportation & Bus Station & 113.67985  & 22.620516 & Nansha District \\ 
19th Chong Station & Transportation & Bus Station & 113.64602  & 22.600134 & Nansha District \\ 
Guangzhou Nansha Binhai Wetland Park & Tourist Attraction & Other & 113.639675 & 22.610122 & Nansha District \\ 
Lotus Pond & Tourist Attraction & Other & 113.644513 & 22.612291 & Nansha District \\ 
Nansha Wetland Scenic Area & Tourist Attraction & Attraction & 113.637737 & 22.60104  & Nansha District \\ 
Logistics Base Phase III Station & Transportation & Bus Station & 113.679959 & 22.620209 & Nansha District \\ 
Nansha Phase III Wharf Station & Transportation & Bus Station & 113.679834 & 22.620475 & Nansha District \\ 
Nansha Wetland Park Station & Transportation & Bus Station & 113.639274 & 22.609453 & Nansha District \\ 
Nansha Wetland School & Culture and Education & Other & 113.653427 & 22.614971 & Nansha District \\ 
Dedicated Parking Lot & Transportation & Parking Lot & 113.637919 & 22.610785 & Nansha District \\ 
19th Chong Aquatic Products Trading Wharf & Shopping & Market & 113.649883 & 22.603448 & Nansha District \\ 
Nansha Waterbird World Ecological Park & Tourist Attraction & Attraction & 113.636382 & 22.615331 & Nansha District \\ 
Nansha 19th Chong Fisherman's Wharf & Transportation & Port & 113.646167 & 22.600382 & Nansha District \\ 
Public Toilet & Living Services & Public Toilet & 113.639733 & 22.609354 & Nansha District \\ 
Public Toilet & Living Services & Public Toilet & 113.642694 & 22.612888 & Nansha District \\ 
Public Toilet & Living Services & Public Toilet & 113.636472 & 22.595375 & Nansha District \\ 
Public Toilet & Living Services & Public Toilet & 113.630324 & 22.590145 & Nansha District \\ 
Qin's Inn · Yi Jiangnan Store & Hospitality & Budget Hotel & 113.649073 & 22.602917 & Nansha District \\ 
Public Toilet & Living Services & Public Toilet & 113.648939 & 22.602766 & Nansha District \\ 
Public Toilet & Living Services & Public Toilet & 113.632644 & 22.592318 & Nansha District \\ 
Nansha Wetland Park & Tourist Attraction & Other & 113.639689 & 22.610232 & Nansha District \\ 
Viewing Fish Pavilion & Tourist Attraction & Other & 113.626726 & 22.58858  & Nansha District \\ 
Egret Pavilion & Tourist Attraction & Other & 113.630688 & 22.590393 & Nansha District \\ 
Guangzhou 19th Chong Apartment & Hospitality & Other & 113.64924  & 22.602902 & Nansha District \\ 
Nansha Wetland - Wetland Floating Bridge & Tourist Attraction & Other & 113.631836 & 22.59294  & Nansha District \\ 
Fu Hai Lai Restaurant & Food & Chinese Cuisine & 113.650675 & 22.604097 & Nansha District \\ 
China Nansha Wetland Xiang Yun Sha Intangible Cultural & Tourist Attraction & Plaza & 113.636263 & 22.595823 & Nansha District \\ 
Hao Jing Fishing Village & Food & Chinese Cuisine & 113.650537 & 22.604126 & Nansha District \\ 
Walking Area Ferry Wharf & Transportation & Port & 113.652771 & 22.614251 & Nansha District \\ 
Nansha Fisheries Industrial Park & Business & Industrial Park & 113.627029 & 22.607906 & Nansha District \\ 
19th Chong Homestay & Hospitality & Hotel & 113.64907  & 22.603012 & Nansha District \\ 
Nansha Bee Bar & Food & Other & 113.640099 & 22.610527 & Nansha District \\ 
Blue Sea Fishing Village & Food & Chinese Cuisine & 113.650062 & 22.603637 & Nansha District \\ 
Guangzhou Butter Crab Aquatic Products Co., Ltd. & Food & Chinese Cuisine & 113.648129 & 22.597649 & Nansha District \\ 
Jin He Fishing Village & Food & Other & 113.650071 & 22.603579 & Nansha District \\ 
Liu Fu Yuan Sour & Food & Other & 113.64882  & 22.602429 & Nansha District \\ 
Fishermen's Village Seafood Restaurant & Food & Chinese Cuisine & 113.650482 & 22.603982 & Nansha District \\ 
Lin's Old Rural Snacks & Food & Chinese Cuisine & 113.649188 & 22.60256  & Nansha District \\ 
Peach Blossom Brew Nansha Branch & Shopping & Other & 113.64933  & 22.603075 & Nansha District \\ 
Authentic Chaoshan Hand-beaten Beef Balls & Food & Other & 113.649274 & 22.602663 & Nansha District \\

\end{longtable}
}
\end{landscape}

\begin{landscape}
{\small\tabcolsep=3pt  % hold it local
\fontsize{9pt}{9pt}\selectfont
\begin{longtable}{c|cccccccc}

    \caption{Frequency Distribution of POIs for 60 Sample Subdistricts in the Pearl River Delta.} \label{tab:poi} \\\hline
    
    \textbf{Subdistrict} & ATM & 3-Star Hotel & Specialized Hosp. & Chinese Cuisine & ... & Residential Area & Conv. Store &  \\
    \midrule
    \endfirsthead

    \multicolumn{6}{c}%
    {{\tablename\ \thetable{} -- continued from previous page}} \\\hline 
    \textbf{Subdistrict} & ATM & 3-Star Hotel & Specialized Hosp. & Chinese Cuisine & ... & Residential Area & Conv. Store &  \\
    \midrule
    \endhead

    \hline \multicolumn{6}{r}{{Continued on next page}} \\ \hline
    \endfoot
    \hline \hline
    \endlastfoot

    Di Dong & 18 & 0 & 12 & 298 & ... & 42 & 120 &  \\
    Feng Yuan & 11 & 3 & 7 & 127 & ... & 22 & 41 &  \\
    Duo Bao & 1 & 0 & 5 & 38 & ... & 13 & 24 &  \\
    Wan Qing Sha & 4 & 0 & 1 & 131 & ... & 7 & 84 &  \\
    Nan Sha & 46 & 2 & 37 & 1273 & ... & 172 & 601 &  \\
    Hai Zhuang & 15 & 5 & 10 & 172 & ... & 50 & 113 &  \\
    Chao Lian & 9 & 0 & 2 & 120 & ... & 18 & 64 &  \\
    Long Kou & 5 & 0 & 5 & 86 & ... & 6 & 51 &  \\
    Sha Ping & 44 & 1 & 30 & 953 & ... & 182 & 627 &  \\
    Cui Xiang & 24 & 1 & 32 & 374 & ... & 116 & 115 &  \\
    Bin Jiang & 7 & 1 & 17 & 150 & ... & 46 & 68 &  \\
    Jiang Nan Zhong & 9 & 0 & 8 & 238 & ... & 89 & 81 &  \\
    Hua Long & 6 & 0 & 1 & 329 & ... & 20 & 189 &  \\
    Nan Cun & 46 & 5 & 74 & 1752 & ... & 178 & 696 &  \\
    Xin Zao & 6 & 0 & 2 & 184 & ... & 18 & 41 &  \\
    Yun Yong Forest Farm & 0 & 0 & 0 & 0 & ... & 0 & 0 &  \\
    Zhong Cun & 27 & 2 & 23 & 845 & ... & 71 & 360 &  \\
    Foshan Prison & 0 & 0 & 0 & 0 & ... & 0 & 0 &  \\
    Qian Jin & 11 & 1 & 32 & 656 & ... & 85 & 275 &  \\
    Lin He & 31 & 6 & 36 & 322 & ... & 62 & 107 &  \\
    Sha He & 6 & 0 & 0 & 59 & ... & 28 & 35 &  \\
    Lan He & 12 & 0 & 4 & 228 & ... & 22 & 168 &  \\
    Hai Long & 6 & 0 & 2 & 267 & ... & 28 & 166 &  \\
    Da Tang & 27 & 1 & 19 & 148 & ... & 64 & 51 &  \\
    Huang Cun & 8 & 2 & 8 & 259 & ... & 30 & 91 &  \\
    San Jiao & 18 & 1 & 10 & 370 & ... & 54 & 301 &  \\
    Chang Gang & 26 & 1 & 20 & 405 & ... & 82 & 114 &  \\
    Tang Xia & 76 & 1 & 79 & 3071 & ... & 375 & 2104 &  \\
    Da Lang & 52 & 6 & 91 & 3089 & ... & 330 & 2034 &  \\
    Qiao Tou & 24 & 2 & 27 & 1244 & ... & 90 & 884 &  \\
    Heng Li & 29 & 0 & 28 & 1280 & ... & 131 & 1004 &  \\
    Dong Keng & 22 & 2 & 20 & 907 & ... & 81 & 593 &  \\
    Zhong Tang & 17 & 2 & 25 & 796 & ... & 64 & 708 &  \\
    Hou Jie & 68 & 4 & 122 & 3106 & ... & 391 & 2219 &  \\
    Sha Tian & 19 & 0 & 18 & 632 & ... & 73 & 519 &  \\
    Hong Mei & 8 & 2 & 4 & 212 & ... & 19 & 165 &  \\
    Pa Zhou & 14 & 2 & 10 & 322 & ... & 50 & 113 &  \\
    Nan Hua West & 2 & 3 & 2 & 20 & ... & 8 & 18 &  \\
    Shi Pai & 21 & 3 & 34 & 1224 & ... & 127 & 1007 &  \\
    Shi Ji & 29 & 2 & 55 & 1303 & ... & 128 & 994 &  \\
    Shi Long & 28 & 2 & 49 & 798 & ... & 90 & 405 &  \\
    Dongguan Port Mgmt. Comm. & 0 & 0 & 0 & 0 & ... & 1 & 1 &  \\
    Ma Chong & 14 & 3 & 13 & 656 & ... & 69 & 393 &  \\
    Cha Shan & 19 & 3 & 21 & 1066 & ... & 124 & 827 &  \\
    Lie De & 53 & 2 & 64 & 453 & ... & 85 & 109 &  \\
    Dao Jiao & 16 & 0 & 15 & 586 & ... & 61 & 458 &  \\
    Gao Bu & 17 & 1 & 24 & 759 & ... & 53 & 629 &  \\
    Wan Jiang & 24 & 3 & 43 & 1237 & ... & 176 & 886 &  \\
    Dong Cheng & 116 & 3 & 189 & 3576 & ... & 595 & 1966 &  \\
    Nan Cheng & 77 & 2 & 127 & 2228 & ... & 337 & 1020 &  \\
    Da Ling Shan & 35 & 3 & 52 & 1698 & ... & 151 & 1201 &  \\
    Wang Niu Dun & 11 & 0 & 7 & 270 & ... & 30 & 223 &  \\
    Dongguan Eco-Park & 2 & 0 & 0 & 75 & ... & 7 & 89 &  \\
    Guan Cheng & 42 & 2 & 44 & 790 & ... & 240 & 502 &  \\
    Song Shan Lake Mgmt. Comm. & 27 & 1 & 15 & 654 & ... & 100 & 403 &  \\
    Dong Feng & 29 & 1 & 38 & 1017 & ... & 90 & 713 &  \\
    Nan Tou & 14 & 0 & 13 & 545 & ... & 66 & 379 &  \\
    Nan Lang & 20 & 0 & 12 & 494 & ... & 96 & 266 &  \\
    Ren Min & 8 & 5 & 11 & 286 & ... & 82 & 130 &  \\
    Dong Sheng & 36 & 1 & 23 & 1212 & ... & 113 & 906 &  \\
\end{longtable}
}
\end{landscape}

\begin{table}[!htbp]
    \caption{Statistics of multi-source data on Beijing and Chengdu.}\label{tab:bjcd}
    \fontsize{10pt}{10pt}\selectfont
    \resizebox{\linewidth}{!}{
        \begin{tabular}{c|cccc}
        \toprule
         \textbf{City} & \textbf{Area} & \textbf{\#POIs} & \textbf{\#SVIs} & \textbf{\#OD-records}\\
        \midrule
         Beijing & 16410 km²& 829802 & 48233 & 80699731\\
         Chengdu & 14335 km² & 708481 & 19980 & 20241895\\
        \bottomrule
        \end{tabular}
    }
\end{table}

% Figure

\begin{figure}[!htbp]
	\centering
		\includegraphics[width=\textwidth]{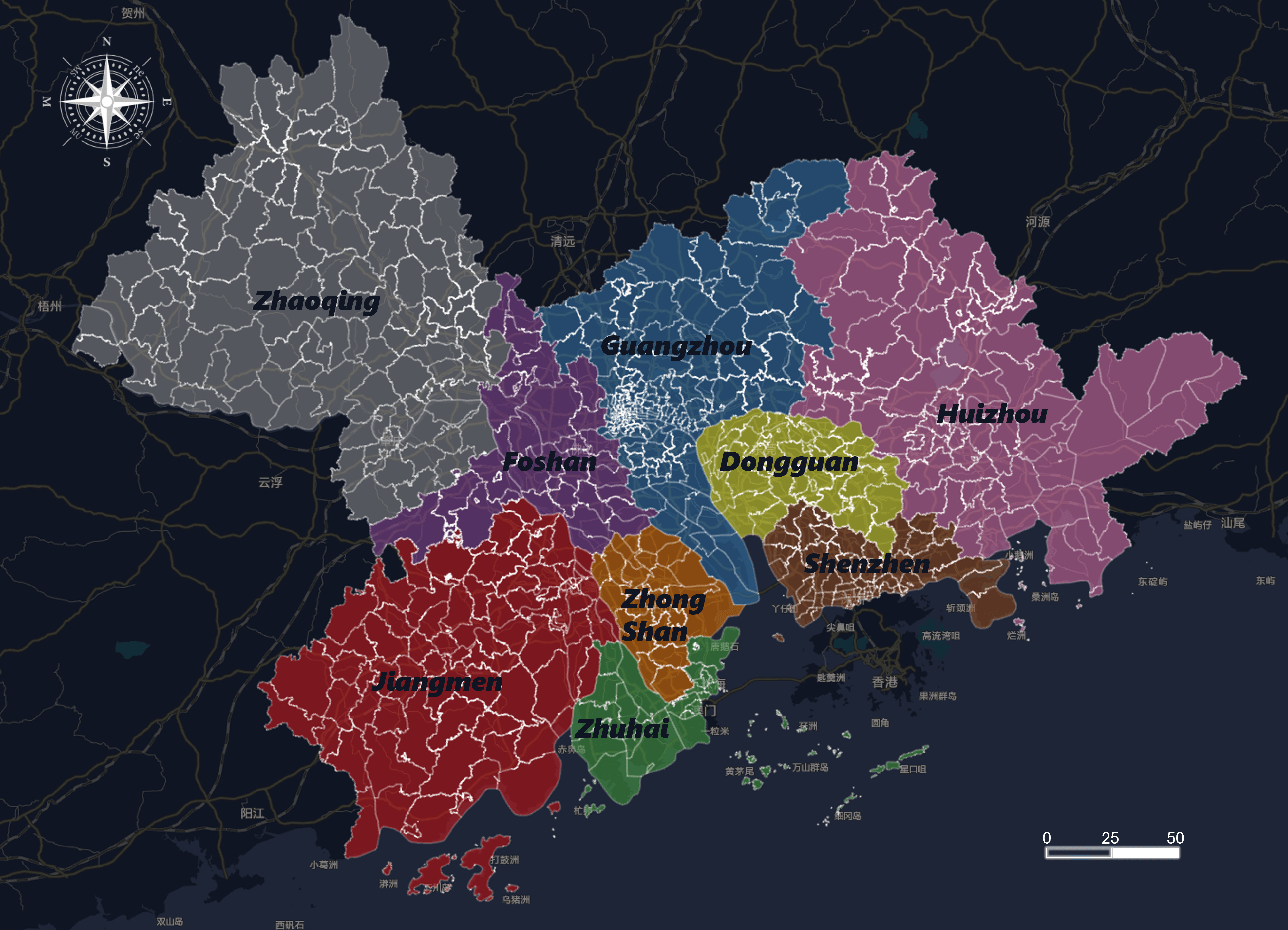}
	  \caption{Administrative Division Schematic of the Pearl River Delta}\label{fig:division}
\end{figure}

\begin{figure}[!htbp]
	\centering
		\includegraphics[width=\textwidth]{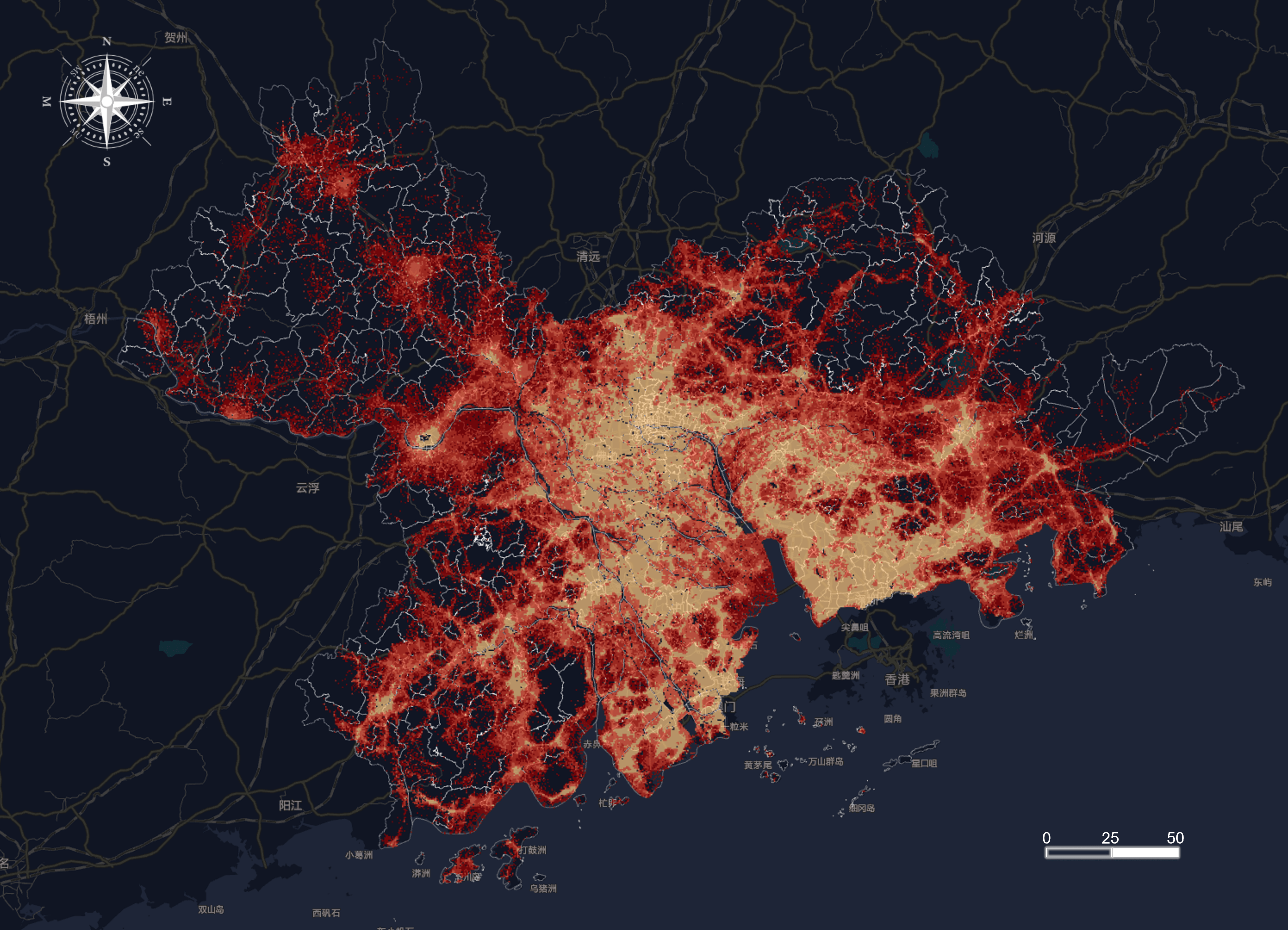}
	  \caption{Spatial Heatmap of Origin Frequencies for Human Mobility in the Pearl River Delta}\label{fig:oflows}
\end{figure}

\begin{figure}[!htbp]
	\centering
		\includegraphics[width=\textwidth]{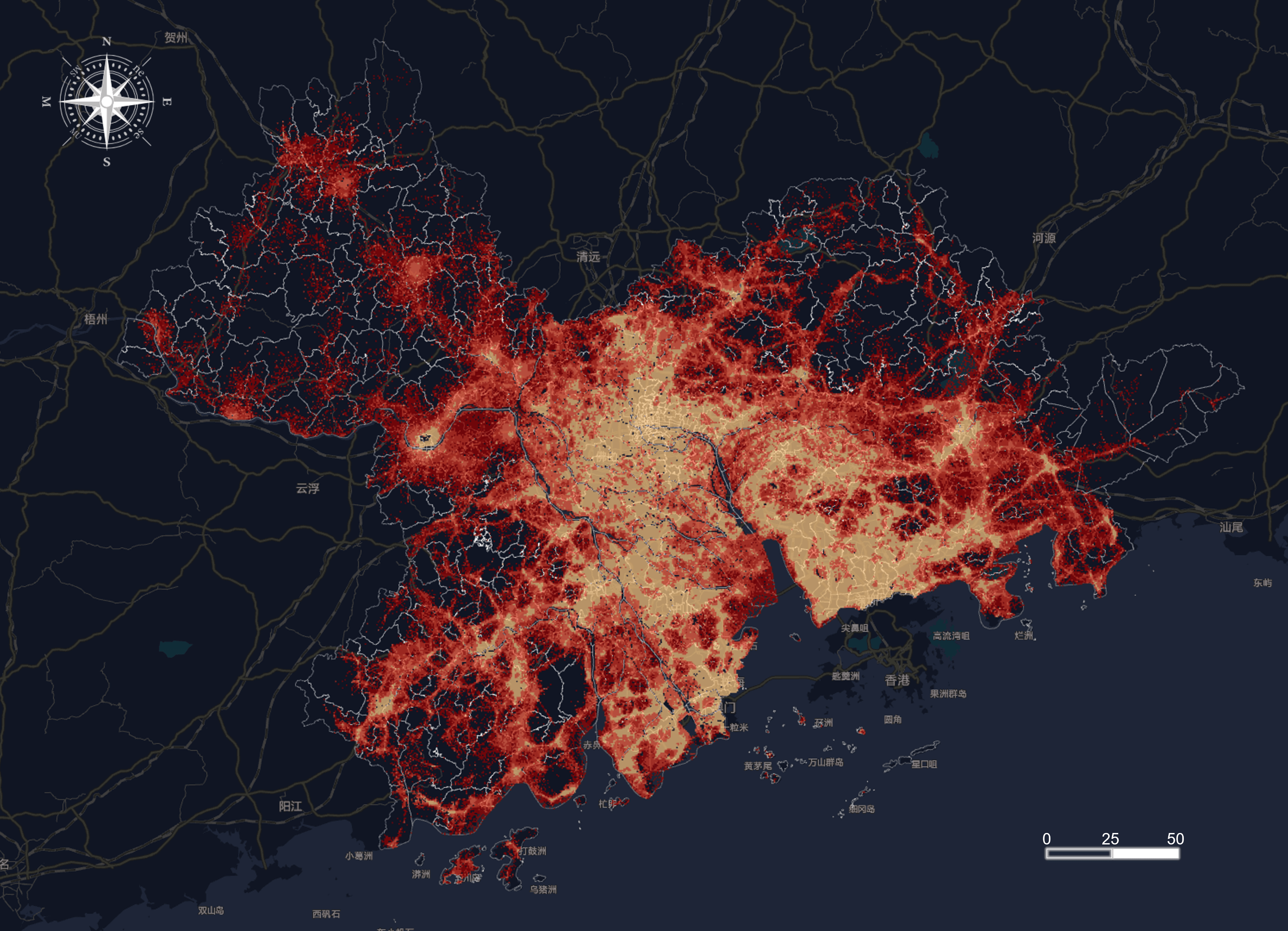}
	  \caption{Spatial Heatmap of Destination Frequencies for Human Mobility in the Pearl River Delta}\label{fig:dflows}
\end{figure}

\newpage
\section{City Group Statistics}\label{app3}

\setcounter{table}{0}

\begin{table}[!htbp]
    \centering
    \caption{GDP Statistics Across Mega, Large, and Major Cities}
    \fontsize{10pt}{10pt}\selectfont
    \resizebox{\linewidth}{!}{
    \begin{tabular}{c|cccc|cccc|cccc}
    \toprule[1.0pt] %\hline
        \multirow{1}{*}{} &\multicolumn{4}{c}{\textbf{Mega Cities}} &\multicolumn{4}{c}{\textbf{Large Cities}} &\multicolumn{4}{c}{\textbf{Major Cities}}\\
        \cmidrule{2-5}\cmidrule{6-9}\cmidrule{10-13}
        & Mean. & Std. & Max. & Min. & Mean. & Std. & Max. & Min. & Mean. & Std. & Max. & Min. \\\hline
        Primary & 17.43 & 23.61 & 72.28 & 0.00 & 49.86 & 25.57 & 88.50 & 0.64 & 50.57 & 37.84 & 119.81 & 0.26\\
        Secondary & 1275.78 & 1483.24 & 6513.64 & 126.05 & 1014.97 & 708.10 & 2478.63 & 83.61 & 339.68 & 408.47 & 1795.25 & 43.68\\
        Tertiary & 2151.68 & 1699.90 & 5765.81 & 245.09 & 745.14 & 590.42 & 1658.26 & 81.61 & 354.76 & 491.54 & 1746.83 & 50.68\\
    \toprule[1.0pt] %\hline
    \end{tabular}
    }
    \label{tab:stgdp}
\end{table}

\begin{table}[!htbp]
    \centering
    \caption{Absolute Error Statistics Across Mega, Cities, and Major Cities}
    \fontsize{10pt}{10pt}\selectfont
    \resizebox{\linewidth}{!}{
    \begin{tabular}{c|cccc|cccc|cccc}
    \toprule[1.0pt] %\hline
        \multirow{1}{*}{} &\multicolumn{4}{c}{\textbf{Mega Cities}} &\multicolumn{4}{c}{\textbf{Large Cities}} &\multicolumn{4}{c}{\textbf{Major Cities}}\\
        \cmidrule{2-5}\cmidrule{6-9}\cmidrule{10-13}
        & Mean. & Std. & Max. & Min. & Mean. & Std. & Max. & Min. & Mean. & Std. & Max. & Min. \\\hline
        Primary & 5.24 & 5.66 & 25.51 & 0.32 &6.59 & 4.69 & 17.98 & 0.37 & 6.57 & 5.33 & 17.04 & 0.39\\
        Secondary & 154.00 & 122.64 & 459.63 & 17.24 & 67.59 & 57.68 & 194.45 & 5.83 & 104.98 & 91.79 & 372.06 & 18.36\\
        Tertiary & 200.43 & 160.72 & 528.12 & 8.61 & 189.89 & 152.39 & 429.07 & 16.57 & 141.81 & 76.39 & 383.27 & 54.67\\
    \toprule[1.0pt] %\hline
    \end{tabular}
    }
    \label{tab:sterr}
\end{table}

\section*{CRediT authorship contribution statement}
\textbf{Jinzhou Cao}: Conceptualization, Investigation, Data curation, Resources, Funding acquisition, Writing - original draft, Writing – review \& editing. 
\textbf{Xiangxu Wang}: Methodology, Formal analysis, Visualization, Writing - original draft,  Writing – review \& editing. 
\textbf{Jiashi Chen}: Writing – review \& editing. 
\textbf{Wei Tu}: Resources, Writing – review \& editing. 
\textbf{Zhenhui Li}: Writing – review \& editing. 
\textbf{Xindong Yang}: Writing – review \& editing. 
\textbf{Tianhong Zhao}: Investigation, Writing – review \& editing. 
\textbf{Qingquan Li}: Resources, Writing – review \& editing.

\section*{Acknowledgements}
This research was supported by Shenzhen Science and Technology Program [grant numbers JCYJ20240813113300001, JYCJ20220530152817039, 20231127180406001, JCYJ
20220818100200001]; the National Natural Science Foundation of China [grant number 42001393]; and the Innovation Team of the Department of Education of Guangdong Province: Unmanned Autonomous Surveying and Mapping [grant number 2024KCXTD013].

\bibliographystyle{elsarticle-harv} 
\bibliography{cas-refs}

\begin{thebibliography}{83}
\expandafter\ifx\csname natexlab\endcsname\relax\def\natexlab#1{#1}\fi
\providecommand{\url}[1]{\texttt{#1}}
\providecommand{\href}[2]{#2}
\providecommand{\path}[1]{#1}
\providecommand{\DOIprefix}{doi:}
\providecommand{\ArXivprefix}{arXiv:}
\providecommand{\URLprefix}{URL: }
\providecommand{\Pubmedprefix}{pmid:}
\providecommand{\doi}[1]{\href{http://dx.doi.org/#1}{\path{#1}}}
\providecommand{\Pubmed}[1]{\href{pmid:#1}{\path{#1}}}
\providecommand{\bibinfo}[2]{#2}
\ifx\xfnm\relax \def\xfnm[#1]{\unskip,\space#1}\fi
%Type = Inproceedings
\bibitem[{Abu-El-Haija et~al.(2018)Abu-El-Haija, Perozzi, Al-Rfou and Alemi}]{abu2018watch}
\bibinfo{author}{Abu-El-Haija, S.}, \bibinfo{author}{Perozzi, B.}, \bibinfo{author}{Al-Rfou, R.}, \bibinfo{author}{Alemi, A.}, \bibinfo{year}{2018}.
\newblock \bibinfo{title}{Watch your step: learning node embeddings via graph attention}, in: \bibinfo{booktitle}{Proceedings of the 32nd International Conference on Neural Information Processing Systems}, \bibinfo{publisher}{Curran Associates Inc.}, \bibinfo{address}{Red Hook, NY, USA}. p. \bibinfo{pages}{9198–9208}.
%Type = Inbook
\bibitem[{Anand and Batra(2022)}]{anand2022application}
\bibinfo{author}{Anand, A.}, \bibinfo{author}{Batra, G.}, \bibinfo{year}{2022}.
\newblock \bibinfo{title}{Application of Geospatial Methods in Evaluating Environmental Interventions and Related Socioeconomic Benefits}. \bibinfo{publisher}{Springer International Publishing}, \bibinfo{address}{Cham}.
\newblock pp. \bibinfo{pages}{275--289}.
%Type = Article
\bibitem[{Bassi et~al.(2024)Bassi, Dertkigil and Cavalli}]{bassi2024improving}
\bibinfo{author}{Bassi, P.R.}, \bibinfo{author}{Dertkigil, S.S.}, \bibinfo{author}{Cavalli, A.}, \bibinfo{year}{2024}.
\newblock \bibinfo{title}{Improving deep neural network generalization and robustness to background bias via layer-wise relevance propagation optimization}.
\newblock \bibinfo{journal}{Nature Communications} \bibinfo{volume}{15}, \bibinfo{pages}{291}.
%Type = Inproceedings
\bibitem[{Belkin and Niyogi(2001)}]{belkin2001laplacian}
\bibinfo{author}{Belkin, M.}, \bibinfo{author}{Niyogi, P.}, \bibinfo{year}{2001}.
\newblock \bibinfo{title}{Laplacian eigenmaps and spectral techniques for embedding and clustering}, in: \bibinfo{booktitle}{Proceedings of the 15th International Conference on Neural Information Processing Systems: Natural and Synthetic}, \bibinfo{publisher}{MIT Press}, \bibinfo{address}{Cambridge, MA, USA}. p. \bibinfo{pages}{585–591}.
%Type = Inbook
\bibitem[{Berthelot et~al.(2019)Berthelot, Carlini, Goodfellow, Oliver, Papernot and Raffel}]{berthelot2019mixmatch}
\bibinfo{author}{Berthelot, D.}, \bibinfo{author}{Carlini, N.}, \bibinfo{author}{Goodfellow, I.}, \bibinfo{author}{Oliver, A.}, \bibinfo{author}{Papernot, N.}, \bibinfo{author}{Raffel, C.}, \bibinfo{year}{2019}.
\newblock \bibinfo{title}{MixMatch: a holistic approach to semi-supervised learning}. \bibinfo{publisher}{Curran Associates Inc.}, \bibinfo{address}{Red Hook, NY, USA}.
%Type = Article
\bibitem[{Cai et~al.(2024)Cai, Song, Zhang, Hu and Jing}]{cai2024local}
\bibinfo{author}{Cai, Z.}, \bibinfo{author}{Song, J.}, \bibinfo{author}{Zhang, T.}, \bibinfo{author}{Hu, C.}, \bibinfo{author}{Jing, X.Y.}, \bibinfo{year}{2024}.
\newblock \bibinfo{title}{Local weight coupled network: Multi-modal unequal semi-supervised domain adaptation}.
\newblock \bibinfo{journal}{Multimedia Tools and Applications} \bibinfo{volume}{83}, \bibinfo{pages}{4331--4357}.
%Type = Article
\bibitem[{Cao et~al.(2025a)Cao, Cao, Tu, Tan, Wang, Chen, Zhang and Li}]{cao2025nighttime}
\bibinfo{author}{Cao, J.}, \bibinfo{author}{Cao, X.}, \bibinfo{author}{Tu, W.}, \bibinfo{author}{Tan, X.}, \bibinfo{author}{Wang, T.}, \bibinfo{author}{Chen, G.}, \bibinfo{author}{Zhang, X.}, \bibinfo{author}{Li, Q.}, \bibinfo{year}{2025}a.
\newblock \bibinfo{title}{Nighttime light imagery or mobile phone footprints: Which better reflects urban socio-economics at the grid level? a case study in the pearl river delta, china}.
\newblock \bibinfo{journal}{Computers, Environment and Urban Systems} \bibinfo{volume}{116}, \bibinfo{pages}{102220}.
%Type = Article
\bibitem[{Cao et~al.(2021)Cao, Li, Tu, Gao, Cao and Zhong}]{cao_ResolvingUrbanMobility_2021}
\bibinfo{author}{Cao, J.}, \bibinfo{author}{Li, Q.}, \bibinfo{author}{Tu, W.}, \bibinfo{author}{Gao, Q.}, \bibinfo{author}{Cao, R.}, \bibinfo{author}{Zhong, C.}, \bibinfo{year}{2021}.
\newblock \bibinfo{title}{Resolving urban mobility networks from individual travel graphs using massive-scale mobile phone tracking data}.
\newblock \bibinfo{journal}{Cities} \bibinfo{volume}{110}, \bibinfo{pages}{103077}.
\newblock \DOIprefix\doi{10.1016/j.cities.2020.103077}.
%Type = Article
\bibitem[{Cao et~al.(2024)Cao, Tu, Cao, Gao, Chen and Li}]{cao_UntanglingAssociationUrban_2024}
\bibinfo{author}{Cao, J.}, \bibinfo{author}{Tu, W.}, \bibinfo{author}{Cao, R.}, \bibinfo{author}{Gao, Q.}, \bibinfo{author}{Chen, G.}, \bibinfo{author}{Li, Q.}, \bibinfo{year}{2024}.
\newblock \bibinfo{title}{Untangling the association between urban mobility and urban elements}.
\newblock \bibinfo{journal}{Geo-spatial Information Science} \bibinfo{volume}{27}, \bibinfo{pages}{1071--1089}.
%Type = Article
\bibitem[{Cao et~al.(2025b)Cao, Wang, Chen, Tu, Shen, Zhao, Chen and Li}]{cao_DisentanglingHourlyDynamics_2025}
\bibinfo{author}{Cao, J.}, \bibinfo{author}{Wang, X.}, \bibinfo{author}{Chen, G.}, \bibinfo{author}{Tu, W.}, \bibinfo{author}{Shen, X.}, \bibinfo{author}{Zhao, T.}, \bibinfo{author}{Chen, J.}, \bibinfo{author}{Li, Q.}, \bibinfo{year}{2025}b.
\newblock \bibinfo{title}{Disentangling the hourly dynamics of mixed urban function: {{A}} multimodal fusion perspective using dynamic graphs}.
\newblock \bibinfo{journal}{Information Fusion} \bibinfo{volume}{117}, \bibinfo{pages}{102832}.
%Type = Article
\bibitem[{Cao et~al.(2022)Cao, Tu, Cai, Zhao, Xiao, Cao, Gao and Su}]{cao2022machine}
\bibinfo{author}{Cao, R.}, \bibinfo{author}{Tu, W.}, \bibinfo{author}{Cai, J.}, \bibinfo{author}{Zhao, T.}, \bibinfo{author}{Xiao, J.}, \bibinfo{author}{Cao, J.}, \bibinfo{author}{Gao, Q.}, \bibinfo{author}{Su, H.}, \bibinfo{year}{2022}.
\newblock \bibinfo{title}{Machine learning-based economic development mapping from multi-source open geospatial data}.
\newblock \bibinfo{journal}{ISPRS Annals of the Photogrammetry, Remote Sensing and Spatial Information Sciences} \bibinfo{volume}{4}, \bibinfo{pages}{259--266}.
%Type = Inproceedings
\bibitem[{Chang et~al.(2020)Chang, Jang, Kim and Kang}]{chang2020learning}
\bibinfo{author}{Chang, B.}, \bibinfo{author}{Jang, G.}, \bibinfo{author}{Kim, S.}, \bibinfo{author}{Kang, J.}, \bibinfo{year}{2020}.
\newblock \bibinfo{title}{Learning graph-based geographical latent representation for point-of-interest recommendation}, in: \bibinfo{booktitle}{Proceedings of the 29th ACM International conference on information \& knowledge management}, pp. \bibinfo{pages}{135--144}.
%Type = Article
\bibitem[{Choromanski et~al.(2020)Choromanski, Likhosherstov, Dohan, Song, Gane, Sarlos, Hawkins, Davis, Mohiuddin, Kaiser et~al.}]{choromanski2020rethinking}
\bibinfo{author}{Choromanski, K.}, \bibinfo{author}{Likhosherstov, V.}, \bibinfo{author}{Dohan, D.}, \bibinfo{author}{Song, X.}, \bibinfo{author}{Gane, A.}, \bibinfo{author}{Sarlos, T.}, \bibinfo{author}{Hawkins, P.}, \bibinfo{author}{Davis, J.}, \bibinfo{author}{Mohiuddin, A.}, \bibinfo{author}{Kaiser, L.}, et~al., \bibinfo{year}{2020}.
\newblock \bibinfo{title}{Rethinking attention with performers}.
\newblock \bibinfo{journal}{arXiv preprint arXiv:2009.14794} .
%Type = Article
\bibitem[{Cui et~al.(2018)Cui, Wang, Pei and Zhu}]{cui2018survey}
\bibinfo{author}{Cui, P.}, \bibinfo{author}{Wang, X.}, \bibinfo{author}{Pei, J.}, \bibinfo{author}{Zhu, W.}, \bibinfo{year}{2018}.
\newblock \bibinfo{title}{A survey on network embedding}.
\newblock \bibinfo{journal}{IEEE transactions on knowledge and data engineering} \bibinfo{volume}{31}, \bibinfo{pages}{833--852}.
%Type = Article
\bibitem[{Custodio et~al.(2023)Custodio, Hadjikakou and Bryan}]{custodio2023review}
\bibinfo{author}{Custodio, H.M.}, \bibinfo{author}{Hadjikakou, M.}, \bibinfo{author}{Bryan, B.A.}, \bibinfo{year}{2023}.
\newblock \bibinfo{title}{A review of socioeconomic indicators of sustainability and wellbeing building on the social foundations framework}.
\newblock \bibinfo{journal}{Ecological Economics} \bibinfo{volume}{203}, \bibinfo{pages}{107608}.
%Type = Article
\bibitem[{Deng et~al.(2023)Deng, Guan, Cai, Yang, Fraedrich, Zhang, Tang, Liao, Wei and Guo}]{deng2023supervised}
\bibinfo{author}{Deng, R.}, \bibinfo{author}{Guan, Y.}, \bibinfo{author}{Cai, D.}, \bibinfo{author}{Yang, T.}, \bibinfo{author}{Fraedrich, K.}, \bibinfo{author}{Zhang, C.}, \bibinfo{author}{Tang, J.}, \bibinfo{author}{Liao, Z.}, \bibinfo{author}{Wei, Z.}, \bibinfo{author}{Guo, S.}, \bibinfo{year}{2023}.
\newblock \bibinfo{title}{Supervised versus semi-supervised urban functional area prediction: uncertainty, robustness and sensitivity}.
\newblock \bibinfo{journal}{Remote Sensing} \bibinfo{volume}{15}, \bibinfo{pages}{341}.
%Type = Article
\bibitem[{Dosovitskiy(2020)}]{dosovitskiy2020image}
\bibinfo{author}{Dosovitskiy, A.}, \bibinfo{year}{2020}.
\newblock \bibinfo{title}{An image is worth 16x16 words: Transformers for image recognition at scale}.
\newblock \bibinfo{journal}{arXiv preprint arXiv:2010.11929} .
%Type = Article
\bibitem[{Dwivedi et~al.(2023)Dwivedi, Joshi, Luu, Laurent, Bengio and Bresson}]{dwivedi2023benchmarking}
\bibinfo{author}{Dwivedi, V.P.}, \bibinfo{author}{Joshi, C.K.}, \bibinfo{author}{Luu, A.T.}, \bibinfo{author}{Laurent, T.}, \bibinfo{author}{Bengio, Y.}, \bibinfo{author}{Bresson, X.}, \bibinfo{year}{2023}.
\newblock \bibinfo{title}{Benchmarking graph neural networks}.
\newblock \bibinfo{journal}{Journal of Machine Learning Research} \bibinfo{volume}{24}, \bibinfo{pages}{1--48}.
%Type = Article
\bibitem[{Dwivedi et~al.(2021)Dwivedi, Luu, Laurent, Bengio and Bresson}]{dwivedi2021graph}
\bibinfo{author}{Dwivedi, V.P.}, \bibinfo{author}{Luu, A.T.}, \bibinfo{author}{Laurent, T.}, \bibinfo{author}{Bengio, Y.}, \bibinfo{author}{Bresson, X.}, \bibinfo{year}{2021}.
\newblock \bibinfo{title}{Graph neural networks with learnable structural and positional representations}.
\newblock \bibinfo{journal}{arXiv preprint arXiv:2110.07875} .
%Type = Article
\bibitem[{Fan et~al.(2021)Fan, Sun, Yao, Zhang, Yan and Sun}]{fan2021well}
\bibinfo{author}{Fan, D.}, \bibinfo{author}{Sun, H.}, \bibinfo{author}{Yao, J.}, \bibinfo{author}{Zhang, K.}, \bibinfo{author}{Yan, X.}, \bibinfo{author}{Sun, Z.}, \bibinfo{year}{2021}.
\newblock \bibinfo{title}{Well production forecasting based on arima-lstm model considering manual operations}.
\newblock \bibinfo{journal}{Energy} \bibinfo{volume}{220}, \bibinfo{pages}{119708}.
%Type = Article
\bibitem[{Fan and Sun(2021)}]{fan2021cpi}
\bibinfo{author}{Fan, Y.}, \bibinfo{author}{Sun, Z.}, \bibinfo{year}{2021}.
\newblock \bibinfo{title}{Cpi big data prediction based on wavelet twin support vector machine}.
\newblock \bibinfo{journal}{International Journal of Pattern Recognition and Artificial Intelligence} \bibinfo{volume}{35}, \bibinfo{pages}{2159013}.
%Type = Article
\bibitem[{Fan et~al.(2023)Fan, Zhang, Loo and Ratti}]{fan2023urban}
\bibinfo{author}{Fan, Z.}, \bibinfo{author}{Zhang, F.}, \bibinfo{author}{Loo, B.P.}, \bibinfo{author}{Ratti, C.}, \bibinfo{year}{2023}.
\newblock \bibinfo{title}{Urban visual intelligence: Uncovering hidden city profiles with street view images}.
\newblock \bibinfo{journal}{Proceedings of the National Academy of Sciences} \bibinfo{volume}{120}, \bibinfo{pages}{e2220417120}.
%Type = Inproceedings
\bibitem[{Grover and Leskovec(2016)}]{grover2016node2vec}
\bibinfo{author}{Grover, A.}, \bibinfo{author}{Leskovec, J.}, \bibinfo{year}{2016}.
\newblock \bibinfo{title}{node2vec: Scalable feature learning for networks}, in: \bibinfo{booktitle}{Proceedings of the 22nd ACM SIGKDD international conference on Knowledge discovery and data mining}, pp. \bibinfo{pages}{855--864}.
%Type = Article
\bibitem[{Jia et~al.(2020)Jia, Jing, Zhu, Chen, Du, Cai, He and Yue}]{jia2020semi}
\bibinfo{author}{Jia, X.}, \bibinfo{author}{Jing, X.Y.}, \bibinfo{author}{Zhu, X.}, \bibinfo{author}{Chen, S.}, \bibinfo{author}{Du, B.}, \bibinfo{author}{Cai, Z.}, \bibinfo{author}{He, Z.}, \bibinfo{author}{Yue, D.}, \bibinfo{year}{2020}.
\newblock \bibinfo{title}{Semi-supervised multi-view deep discriminant representation learning}.
\newblock \bibinfo{journal}{IEEE transactions on pattern analysis and machine intelligence} \bibinfo{volume}{43}, \bibinfo{pages}{2496--2509}.
%Type = Inproceedings
\bibitem[{Jin et~al.(2023)Jin, Liu, Li and Huang}]{jin2023spatio}
\bibinfo{author}{Jin, G.}, \bibinfo{author}{Liu, L.}, \bibinfo{author}{Li, F.}, \bibinfo{author}{Huang, J.}, \bibinfo{year}{2023}.
\newblock \bibinfo{title}{Spatio-temporal graph neural point process for traffic congestion event prediction}, in: \bibinfo{booktitle}{Proceedings of the AAAI Conference on Artificial Intelligence}, pp. \bibinfo{pages}{14268--14276}.
%Type = Article
\bibitem[{Kim and Swanson(2018)}]{kim2018methods}
\bibinfo{author}{Kim, H.H.}, \bibinfo{author}{Swanson, N.R.}, \bibinfo{year}{2018}.
\newblock \bibinfo{title}{Methods for backcasting, nowcasting and forecasting using factor-midas: With an application to korean gdp}.
\newblock \bibinfo{journal}{Journal of Forecasting} \bibinfo{volume}{37}, \bibinfo{pages}{281--302}.
%Type = Inproceedings
\bibitem[{Kirillov et~al.(2023)Kirillov, Mintun, Ravi, Mao, Rolland, Gustafson, Xiao, Whitehead, Berg, Lo et~al.}]{kirillov2023segment}
\bibinfo{author}{Kirillov, A.}, \bibinfo{author}{Mintun, E.}, \bibinfo{author}{Ravi, N.}, \bibinfo{author}{Mao, H.}, \bibinfo{author}{Rolland, C.}, \bibinfo{author}{Gustafson, L.}, \bibinfo{author}{Xiao, T.}, \bibinfo{author}{Whitehead, S.}, \bibinfo{author}{Berg, A.C.}, \bibinfo{author}{Lo, W.Y.}, et~al., \bibinfo{year}{2023}.
\newblock \bibinfo{title}{Segment anything}, in: \bibinfo{booktitle}{Proceedings of the IEEE/CVF International Conference on Computer Vision}, pp. \bibinfo{pages}{4015--4026}.
%Type = Article
\bibitem[{Laine and Aila(2016)}]{laine2016temporal}
\bibinfo{author}{Laine, S.}, \bibinfo{author}{Aila, T.}, \bibinfo{year}{2016}.
\newblock \bibinfo{title}{Temporal ensembling for semi-supervised learning}.
\newblock \bibinfo{journal}{arXiv preprint arXiv:1610.02242} .
%Type = Inproceedings
\bibitem[{Lee et~al.(2023)Lee, Woo, Moon and Lee}]{lee2023unsupervised}
\bibinfo{author}{Lee, J.}, \bibinfo{author}{Woo, J.O.}, \bibinfo{author}{Moon, H.}, \bibinfo{author}{Lee, K.}, \bibinfo{year}{2023}.
\newblock \bibinfo{title}{Unsupervised accuracy estimation of deep visual models using domain-adaptive adversarial perturbation without source samples}, in: \bibinfo{booktitle}{Proceedings of the IEEE/CVF International Conference on Computer Vision}, pp. \bibinfo{pages}{16443--16452}.
%Type = Inproceedings
\bibitem[{Li et~al.(2022)Li, Xin, Xi, Tarkoma, Hui and Li}]{li2022predicting}
\bibinfo{author}{Li, T.}, \bibinfo{author}{Xin, S.}, \bibinfo{author}{Xi, Y.}, \bibinfo{author}{Tarkoma, S.}, \bibinfo{author}{Hui, P.}, \bibinfo{author}{Li, Y.}, \bibinfo{year}{2022}.
\newblock \bibinfo{title}{Predicting multi-level socioeconomic indicators from structural urban imagery}, in: \bibinfo{booktitle}{Proceedings of the 31st ACM international conference on information \& knowledge management}, pp. \bibinfo{pages}{3282--3291}.
%Type = Inproceedings
\bibitem[{Li et~al.(2023)Li, Huang, Cong, Wang and Wang}]{li2023urban}
\bibinfo{author}{Li, Y.}, \bibinfo{author}{Huang, W.}, \bibinfo{author}{Cong, G.}, \bibinfo{author}{Wang, H.}, \bibinfo{author}{Wang, Z.}, \bibinfo{year}{2023}.
\newblock \bibinfo{title}{Urban region representation learning with openstreetmap building footprints}, in: \bibinfo{booktitle}{Proceedings of the 29th ACM SIGKDD Conference on Knowledge Discovery and Data Mining}, pp. \bibinfo{pages}{1363--1373}.
%Type = Article
\bibitem[{Li et~al.(2017)Li, Yu, Shahabi and Liu}]{li2017diffusion}
\bibinfo{author}{Li, Y.}, \bibinfo{author}{Yu, R.}, \bibinfo{author}{Shahabi, C.}, \bibinfo{author}{Liu, Y.}, \bibinfo{year}{2017}.
\newblock \bibinfo{title}{Diffusion convolutional recurrent neural network: Data-driven traffic forecasting}.
\newblock \bibinfo{journal}{arXiv preprint arXiv:1707.01926} .
%Type = Article
\bibitem[{Liu and Biljecki(2022)}]{liu2022review}
\bibinfo{author}{Liu, P.}, \bibinfo{author}{Biljecki, F.}, \bibinfo{year}{2022}.
\newblock \bibinfo{title}{A review of spatially-explicit geoai applications in urban geography}.
\newblock \bibinfo{journal}{International Journal of Applied Earth Observation and Geoinformation} \bibinfo{volume}{112}, \bibinfo{pages}{102936}.
%Type = Article
\bibitem[{Liu et~al.(2025)Liu, Cheng, Zhang, Xu and Han}]{liu2024capsule}
\bibinfo{author}{Liu, Y.}, \bibinfo{author}{Cheng, D.}, \bibinfo{author}{Zhang, D.}, \bibinfo{author}{Xu, S.}, \bibinfo{author}{Han, J.}, \bibinfo{year}{2025}.
\newblock \bibinfo{title}{Capsule networks with residual pose routing}.
\newblock \bibinfo{journal}{IEEE Transactions on Neural Networks and Learning Systems} \bibinfo{volume}{36}, \bibinfo{pages}{2648--2661}.
%Type = Misc
\bibitem[{Liu et~al.(2024)Liu, Li, Xu and Han}]{liu2025part}
\bibinfo{author}{Liu, Y.}, \bibinfo{author}{Li, C.}, \bibinfo{author}{Xu, S.}, \bibinfo{author}{Han, J.}, \bibinfo{year}{2024}.
\newblock \bibinfo{title}{Part-whole relational fusion towards multi-modal scene understanding}.
\newblock \href{http://arxiv.org/abs/2410.14944}{{\tt arXiv:2410.14944}}.
%Type = Inproceedings
\bibitem[{Lundberg and Lee(2017)}]{lundberg2017unified}
\bibinfo{author}{Lundberg, S.M.}, \bibinfo{author}{Lee, S.I.}, \bibinfo{year}{2017}.
\newblock \bibinfo{title}{A unified approach to interpreting model predictions}, in: \bibinfo{booktitle}{Proceedings of the 31st International Conference on Neural Information Processing Systems}, \bibinfo{publisher}{Curran Associates Inc.}, \bibinfo{address}{Red Hook, NY, USA}. p. \bibinfo{pages}{4768–4777}.
%Type = Inproceedings
\bibitem[{Luo et~al.(2022)Luo, Chung and Chen}]{luo2022urban}
\bibinfo{author}{Luo, Y.}, \bibinfo{author}{Chung, F.l.}, \bibinfo{author}{Chen, K.}, \bibinfo{year}{2022}.
\newblock \bibinfo{title}{Urban region profiling via multi-graph representation learning}, in: \bibinfo{booktitle}{Proceedings of the 31st ACM international conference on information \& knowledge management}, pp. \bibinfo{pages}{4294--4298}.
%Type = Inproceedings
\bibitem[{Ma et~al.(2023)Ma, Karaku{\c{s}} and Rosin}]{ma2023confidence}
\bibinfo{author}{Ma, W.}, \bibinfo{author}{Karaku{\c{s}}, O.}, \bibinfo{author}{Rosin, P.L.}, \bibinfo{year}{2023}.
\newblock \bibinfo{title}{Confidence guided semi-supervised learning in land cover classification}, in: \bibinfo{booktitle}{IGARSS 2023-2023 IEEE International Geoscience and Remote Sensing Symposium}, \bibinfo{organization}{IEEE}. pp. \bibinfo{pages}{5487--5490}.
%Type = Article
\bibitem[{Machicao et~al.(2022)Machicao, Specht, Vellenich, Meneguzzi, David, Stall, Ferraz, Mabile, O'Brien and Corrêa}]{machicao2022deep}
\bibinfo{author}{Machicao, J.}, \bibinfo{author}{Specht, A.}, \bibinfo{author}{Vellenich, D.}, \bibinfo{author}{Meneguzzi, L.}, \bibinfo{author}{David, R.}, \bibinfo{author}{Stall, S.}, \bibinfo{author}{Ferraz, K.}, \bibinfo{author}{Mabile, L.}, \bibinfo{author}{O'Brien, M.}, \bibinfo{author}{Corrêa, P.}, \bibinfo{year}{2022}.
\newblock \bibinfo{title}{A deep-learning method for the prediction of socio-economic indicators from street-view imagery using a case study from brazil}.
\newblock \bibinfo{journal}{Data Science Journal} \bibinfo{volume}{21}.
%Type = Article
\bibitem[{Miao et~al.(2021)Miao, Wang and Li}]{miao2021analyzing}
\bibinfo{author}{Miao, R.}, \bibinfo{author}{Wang, Y.}, \bibinfo{author}{Li, S.}, \bibinfo{year}{2021}.
\newblock \bibinfo{title}{Analyzing urban spatial patterns and functional zones using sina weibo poi data: A case study of beijing}.
\newblock \bibinfo{journal}{Sustainability} \bibinfo{volume}{13}, \bibinfo{pages}{647}.
%Type = Article
\bibitem[{Murugesan et~al.(2022)Murugesan, Mishra and Krishnan}]{murugesan2022forecasting}
\bibinfo{author}{Murugesan, R.}, \bibinfo{author}{Mishra, E.}, \bibinfo{author}{Krishnan, A.H.}, \bibinfo{year}{2022}.
\newblock \bibinfo{title}{Forecasting agricultural commodities prices using deep learning-based models: basic lstm, bi-lstm, stacked lstm, cnn lstm, and convolutional lstm}.
\newblock \bibinfo{journal}{International Journal of Sustainable Agricultural Management and Informatics} \bibinfo{volume}{8}, \bibinfo{pages}{242--277}.
%Type = Inproceedings
\bibitem[{Naaz et~al.(2024)Naaz, Pandey and Lakshmi}]{naaz2024forecasting}
\bibinfo{author}{Naaz, S.}, \bibinfo{author}{Pandey, H.}, \bibinfo{author}{Lakshmi, C.}, \bibinfo{year}{2024}.
\newblock \bibinfo{title}{Forecasting gdp per capita using machine learning algorithms}, in: \bibinfo{booktitle}{2024 Second International Conference on Emerging Trends in Information Technology and Engineering (ICETITE)}, \bibinfo{organization}{IEEE}. pp. \bibinfo{pages}{1--5}.
%Type = Inproceedings
\bibitem[{Pappalardo et~al.(2015)Pappalardo, Pedreschi, Smoreda and Giannotti}]{pappalardo2015using}
\bibinfo{author}{Pappalardo, L.}, \bibinfo{author}{Pedreschi, D.}, \bibinfo{author}{Smoreda, Z.}, \bibinfo{author}{Giannotti, F.}, \bibinfo{year}{2015}.
\newblock \bibinfo{title}{Using big data to study the link between human mobility and socio-economic development}, in: \bibinfo{booktitle}{2015 IEEE international conference on big data (big data)}, \bibinfo{organization}{IEEE}. pp. \bibinfo{pages}{871--878}.
%Type = Inproceedings
\bibitem[{Perozzi et~al.(2014)Perozzi, Al-Rfou and Skiena}]{perozzi2014deepwalk}
\bibinfo{author}{Perozzi, B.}, \bibinfo{author}{Al-Rfou, R.}, \bibinfo{author}{Skiena, S.}, \bibinfo{year}{2014}.
\newblock \bibinfo{title}{Deepwalk: Online learning of social representations}, in: \bibinfo{booktitle}{Proceedings of the 20th ACM SIGKDD international conference on Knowledge discovery and data mining}, pp. \bibinfo{pages}{701--710}.
%Type = Article
\bibitem[{Price and Atkinson(2022)}]{price2022global}
\bibinfo{author}{Price, N.}, \bibinfo{author}{Atkinson, P.M.}, \bibinfo{year}{2022}.
\newblock \bibinfo{title}{Global gdp prediction with night-lights and transfer learning}.
\newblock \bibinfo{journal}{IEEE Journal of Selected Topics in Applied Earth Observations and Remote Sensing} \bibinfo{volume}{15}, \bibinfo{pages}{7128--7138}.
%Type = Article
\bibitem[{Provost and Fawcett(2013)}]{provost2013data}
\bibinfo{author}{Provost, F.}, \bibinfo{author}{Fawcett, T.}, \bibinfo{year}{2013}.
\newblock \bibinfo{title}{Data science and its relationship to big data and data-driven decision making}.
\newblock \bibinfo{journal}{Big data} \bibinfo{volume}{1}, \bibinfo{pages}{51--59}.
%Type = Article
\bibitem[{Puttanapong et~al.(2022)Puttanapong, Martinez~Jr, Bulan, Addawe, Durante and Martillan}]{puttanapong2022predicting}
\bibinfo{author}{Puttanapong, N.}, \bibinfo{author}{Martinez~Jr, A.}, \bibinfo{author}{Bulan, J.A.N.}, \bibinfo{author}{Addawe, M.}, \bibinfo{author}{Durante, R.L.}, \bibinfo{author}{Martillan, M.}, \bibinfo{year}{2022}.
\newblock \bibinfo{title}{Predicting poverty using geospatial data in thailand}.
\newblock \bibinfo{journal}{ISPRS International Journal of Geo-Information} \bibinfo{volume}{11}, \bibinfo{pages}{293}.
%Type = Article
\bibitem[{Qiao et~al.(2024)Qiao, Xiao, Guo, Luo and Xiong}]{qiao2024information}
\bibinfo{author}{Qiao, Z.}, \bibinfo{author}{Xiao, M.}, \bibinfo{author}{Guo, W.}, \bibinfo{author}{Luo, X.}, \bibinfo{author}{Xiong, H.}, \bibinfo{year}{2024}.
\newblock \bibinfo{title}{Information filtering and interpolating for semi-supervised graph domain adaptation}.
\newblock \bibinfo{journal}{Pattern Recognition} \bibinfo{volume}{153}, \bibinfo{pages}{110498}.
%Type = Article
\bibitem[{Richardson et~al.(2021)Richardson, {van Florenstein Mulder} and Vehbi}]{richardson2021nowcasting}
\bibinfo{author}{Richardson, A.}, \bibinfo{author}{{van Florenstein Mulder}, T.}, \bibinfo{author}{Vehbi, T.}, \bibinfo{year}{2021}.
\newblock \bibinfo{title}{Nowcasting gdp using machine-learning algorithms: A real-time assessment}.
\newblock \bibinfo{journal}{International Journal of Forecasting} \bibinfo{volume}{37}, \bibinfo{pages}{941--948}.
%Type = Article
\bibitem[{Roweis and Saul(2000)}]{roweis2000nonlinear}
\bibinfo{author}{Roweis, S.T.}, \bibinfo{author}{Saul, L.K.}, \bibinfo{year}{2000}.
\newblock \bibinfo{title}{Nonlinear dimensionality reduction by locally linear embedding}.
\newblock \bibinfo{journal}{science} \bibinfo{volume}{290}, \bibinfo{pages}{2323--2326}.
%Type = Inproceedings
\bibitem[{Selvaraju et~al.(2017)Selvaraju, Cogswell, Das, Vedantam, Parikh and Batra}]{selvaraju2017grad}
\bibinfo{author}{Selvaraju, R.R.}, \bibinfo{author}{Cogswell, M.}, \bibinfo{author}{Das, A.}, \bibinfo{author}{Vedantam, R.}, \bibinfo{author}{Parikh, D.}, \bibinfo{author}{Batra, D.}, \bibinfo{year}{2017}.
\newblock \bibinfo{title}{Grad-cam: Visual explanations from deep networks via gradient-based localization}, in: \bibinfo{booktitle}{2017 IEEE International Conference on Computer Vision (ICCV)}, pp. \bibinfo{pages}{618--626}.
%Type = Article
\bibitem[{Shi et~al.(2024)Shi, Wang, Guo, Xu, Shen and Cheng}]{shi2024graph}
\bibinfo{author}{Shi, B.}, \bibinfo{author}{Wang, Y.}, \bibinfo{author}{Guo, F.}, \bibinfo{author}{Xu, B.}, \bibinfo{author}{Shen, H.}, \bibinfo{author}{Cheng, X.}, \bibinfo{year}{2024}.
\newblock \bibinfo{title}{Graph domain adaptation: Challenges, progress and prospects}.
\newblock \bibinfo{journal}{arXiv preprint arXiv:2402.00904} .
%Type = Article
\bibitem[{Shi et~al.(2022)Shi, Wu, Li and Li}]{shi2022population}
\bibinfo{author}{Shi, K.}, \bibinfo{author}{Wu, Y.}, \bibinfo{author}{Li, D.}, \bibinfo{author}{Li, X.}, \bibinfo{year}{2022}.
\newblock \bibinfo{title}{Population, gdp, and carbon emissions as revealed by snpp-viirs nighttime light data in china with different scales}.
\newblock \bibinfo{journal}{IEEE Geoscience and Remote Sensing Letters} \bibinfo{volume}{19}, \bibinfo{pages}{1--5}.
%Type = Article
\bibitem[{Siami-Namini and Namin(2018)}]{siami2018forecasting}
\bibinfo{author}{Siami-Namini, S.}, \bibinfo{author}{Namin, A.S.}, \bibinfo{year}{2018}.
\newblock \bibinfo{title}{Forecasting economics and financial time series: Arima vs. lstm}.
\newblock \bibinfo{journal}{arXiv preprint arXiv:1803.06386} .
%Type = Article
\bibitem[{Stock and Watson(2001)}]{stock2001vector}
\bibinfo{author}{Stock, J.H.}, \bibinfo{author}{Watson, M.W.}, \bibinfo{year}{2001}.
\newblock \bibinfo{title}{Vector autoregressions}.
\newblock \bibinfo{journal}{Journal of Economic perspectives} \bibinfo{volume}{15}, \bibinfo{pages}{101--115}.
%Type = Inproceedings
\bibitem[{Sun et~al.(2023)Sun, Shi and Li}]{sun2024graph}
\bibinfo{author}{Sun, Y.}, \bibinfo{author}{Shi, Z.}, \bibinfo{author}{Li, Y.}, \bibinfo{year}{2023}.
\newblock \bibinfo{title}{A graph-theoretic framework for understanding open-world semi-supervised learning}, in: \bibinfo{booktitle}{Proceedings of the 37th International Conference on Neural Information Processing Systems}, \bibinfo{publisher}{Curran Associates Inc.}, \bibinfo{address}{Red Hook, NY, USA}.
%Type = Inproceedings
\bibitem[{Tang et~al.(2015)Tang, Qu, Wang, Zhang, Yan and Mei}]{tang2015line}
\bibinfo{author}{Tang, J.}, \bibinfo{author}{Qu, M.}, \bibinfo{author}{Wang, M.}, \bibinfo{author}{Zhang, M.}, \bibinfo{author}{Yan, J.}, \bibinfo{author}{Mei, Q.}, \bibinfo{year}{2015}.
\newblock \bibinfo{title}{Line: Large-scale information network embedding}, in: \bibinfo{booktitle}{Proceedings of the 24th international conference on world wide web}, pp. \bibinfo{pages}{1067--1077}.
%Type = Inproceedings
\bibitem[{{\"U}lker and {\"U}lker(2019)}]{ulker2019unemployment}
\bibinfo{author}{{\"U}lker, E.D.}, \bibinfo{author}{{\"U}lker, S.}, \bibinfo{year}{2019}.
\newblock \bibinfo{title}{Unemployment rate and gdp prediction using support vector regression}, in: \bibinfo{booktitle}{Proceedings of the 1st International Conference on Advanced Information Science and System}, pp. \bibinfo{pages}{1--5}.
%Type = Article
\bibitem[{Velickovic et~al.(2017)Velickovic, Cucurull, Casanova, Romero, Lio, Bengio et~al.}]{velickovic2017graph}
\bibinfo{author}{Velickovic, P.}, \bibinfo{author}{Cucurull, G.}, \bibinfo{author}{Casanova, A.}, \bibinfo{author}{Romero, A.}, \bibinfo{author}{Lio, P.}, \bibinfo{author}{Bengio, Y.}, et~al., \bibinfo{year}{2017}.
\newblock \bibinfo{title}{Graph attention networks}.
\newblock \bibinfo{journal}{stat} \bibinfo{volume}{1050}, \bibinfo{pages}{10--48550}.
%Type = Article
\bibitem[{Veli{\v{c}}kovi{\'c} et~al.(2018)Veli{\v{c}}kovi{\'c}, Fedus, Hamilton, Li{\`o}, Bengio and Hjelm}]{velivckovic2018deep}
\bibinfo{author}{Veli{\v{c}}kovi{\'c}, P.}, \bibinfo{author}{Fedus, W.}, \bibinfo{author}{Hamilton, W.L.}, \bibinfo{author}{Li{\`o}, P.}, \bibinfo{author}{Bengio, Y.}, \bibinfo{author}{Hjelm, R.D.}, \bibinfo{year}{2018}.
\newblock \bibinfo{title}{Deep graph infomax}.
\newblock \bibinfo{journal}{arXiv preprint arXiv:1809.10341} .
%Type = Article
\bibitem[{Victor-Edema and Essi(2016)}]{victor2016autoregressive}
\bibinfo{author}{Victor-Edema, U.A.}, \bibinfo{author}{Essi, I.D.}, \bibinfo{year}{2016}.
\newblock \bibinfo{title}{Autoregressive integrated moving average with exogenous variable (arimax) model for nigerian non-oil export}.
\newblock \bibinfo{journal}{European Journal of Business and Management} \bibinfo{volume}{8}, \bibinfo{pages}{29--34}.
%Type = Article
\bibitem[{Wang et~al.(2021)Wang, Wang, Xiang, Yu, Deng and Xu}]{wang2021attentive}
\bibinfo{author}{Wang, D.}, \bibinfo{author}{Wang, X.}, \bibinfo{author}{Xiang, Z.}, \bibinfo{author}{Yu, D.}, \bibinfo{author}{Deng, S.}, \bibinfo{author}{Xu, G.}, \bibinfo{year}{2021}.
\newblock \bibinfo{title}{Attentive sequential model based on graph neural network for next poi recommendation}.
\newblock \bibinfo{journal}{World Wide Web} \bibinfo{volume}{24}, \bibinfo{pages}{2161--2184}.
%Type = Inproceedings
\bibitem[{Wang and Li(2017)}]{wang2017region}
\bibinfo{author}{Wang, H.}, \bibinfo{author}{Li, Z.}, \bibinfo{year}{2017}.
\newblock \bibinfo{title}{Region representation learning via mobility flow}, in: \bibinfo{booktitle}{Proceedings of the 2017 ACM on Conference on Information and Knowledge Management}, pp. \bibinfo{pages}{237--246}.
%Type = Inproceedings
\bibitem[{Wang et~al.(2018)Wang, Yang, He, Wang, Zhang and Zhang}]{wang2018urban}
\bibinfo{author}{Wang, W.}, \bibinfo{author}{Yang, S.}, \bibinfo{author}{He, Z.}, \bibinfo{author}{Wang, M.}, \bibinfo{author}{Zhang, J.}, \bibinfo{author}{Zhang, W.}, \bibinfo{year}{2018}.
\newblock \bibinfo{title}{Urban perception of commercial activeness from satellite images and streetscapes}, in: \bibinfo{booktitle}{Companion Proceedings of the The Web Conference 2018}, pp. \bibinfo{pages}{647--654}.
%Type = Article
\bibitem[{Wang et~al.(2023)Wang, Jing, Huang, Jin and Lv}]{wang2023adaptive}
\bibinfo{author}{Wang, Y.}, \bibinfo{author}{Jing, C.}, \bibinfo{author}{Huang, W.}, \bibinfo{author}{Jin, S.}, \bibinfo{author}{Lv, X.}, \bibinfo{year}{2023}.
\newblock \bibinfo{title}{Adaptive spatiotemporal inceptionnet for traffic flow forecasting}.
\newblock \bibinfo{journal}{IEEE Transactions on Intelligent Transportation Systems} \bibinfo{volume}{24}, \bibinfo{pages}{3882--3907}.
%Type = Article
\bibitem[{Wang and Zhu(2024)}]{wang2024hyperGCN}
\bibinfo{author}{Wang, Y.}, \bibinfo{author}{Zhu, D.}, \bibinfo{year}{2024}.
\newblock \bibinfo{title}{A hypergraph-based hybrid graph convolutional network for intracity human activity intensity prediction and geographic relationship interpretation}.
\newblock \bibinfo{journal}{Information Fusion} \bibinfo{volume}{104}, \bibinfo{pages}{102149}.
%Type = Article
\bibitem[{Wang et~al.(2004)Wang, Bovik, Sheikh and Simoncelli}]{wang2004image}
\bibinfo{author}{Wang, Z.}, \bibinfo{author}{Bovik, A.C.}, \bibinfo{author}{Sheikh, H.R.}, \bibinfo{author}{Simoncelli, E.P.}, \bibinfo{year}{2004}.
\newblock \bibinfo{title}{Image quality assessment: from error visibility to structural similarity}.
\newblock \bibinfo{journal}{IEEE transactions on image processing} \bibinfo{volume}{13}, \bibinfo{pages}{600--612}.
%Type = Article
\bibitem[{Wang et~al.(2025)Wang, Zheng, Han, Lu, Yu, Yang and Han}]{wang2024comprehensive}
\bibinfo{author}{Wang, Z.}, \bibinfo{author}{Zheng, J.}, \bibinfo{author}{Han, C.}, \bibinfo{author}{Lu, B.}, \bibinfo{author}{Yu, D.}, \bibinfo{author}{Yang, J.}, \bibinfo{author}{Han, L.}, \bibinfo{year}{2025}.
\newblock \bibinfo{title}{A comprehensive assessment approach for multiscale regional economic development: Fusion modeling of nighttime lights and openstreetmap data}.
\newblock \bibinfo{journal}{Geography and Sustainability} \bibinfo{volume}{6}, \bibinfo{pages}{100230}.
%Type = Incollection
\bibitem[{Wei et~al.(2023)Wei, Meyer and Bacchin}]{wei2023spatial}
\bibinfo{author}{Wei, D.}, \bibinfo{author}{Meyer, H.}, \bibinfo{author}{Bacchin, T.K.}, \bibinfo{year}{2023}.
\newblock \bibinfo{title}{Spatial dynamics in the pearl river delta and development strategies}, in: \bibinfo{booktitle}{Adaptive Urban Transformation: Urban Landscape Dynamics, Regional Design and Territorial Governance in the Pearl River Delta, China}. \bibinfo{publisher}{Springer International Publishing Cham}, pp. \bibinfo{pages}{37--59}.
%Type = Article
\bibitem[{Wu et~al.(2022)Wu, Yan, Fan, Pan, Zhu, Zheng, Cheng and Wang}]{wu2022multi}
\bibinfo{author}{Wu, S.}, \bibinfo{author}{Yan, X.}, \bibinfo{author}{Fan, X.}, \bibinfo{author}{Pan, S.}, \bibinfo{author}{Zhu, S.}, \bibinfo{author}{Zheng, C.}, \bibinfo{author}{Cheng, M.}, \bibinfo{author}{Wang, C.}, \bibinfo{year}{2022}.
\newblock \bibinfo{title}{Multi-graph fusion networks for urban region embedding}.
\newblock \bibinfo{journal}{arXiv preprint arXiv:2201.09760} .
%Type = Article
\bibitem[{Wu et~al.(2021)Wu, Zhang, Chang and Huang}]{wu2021data}
\bibinfo{author}{Wu, X.}, \bibinfo{author}{Zhang, Z.}, \bibinfo{author}{Chang, H.}, \bibinfo{author}{Huang, Q.}, \bibinfo{year}{2021}.
\newblock \bibinfo{title}{A data-driven gross domestic product forecasting model based on multi-indicator assessment}.
\newblock \bibinfo{journal}{Ieee Access} \bibinfo{volume}{9}, \bibinfo{pages}{99495--99503}.
%Type = Article
\bibitem[{Wu et~al.(2019)Wu, Pan, Long, Jiang and Zhang}]{wu2019graph}
\bibinfo{author}{Wu, Z.}, \bibinfo{author}{Pan, S.}, \bibinfo{author}{Long, G.}, \bibinfo{author}{Jiang, J.}, \bibinfo{author}{Zhang, C.}, \bibinfo{year}{2019}.
\newblock \bibinfo{title}{Graph wavenet for deep spatial-temporal graph modeling}.
\newblock \bibinfo{journal}{arXiv preprint arXiv:1906.00121} .
%Type = Article
\bibitem[{Xu et~al.(2024)Xu, Zhang, Zhang, Zhang and Han}]{xu2024deep}
\bibinfo{author}{Xu, C.}, \bibinfo{author}{Zhang, T.}, \bibinfo{author}{Zhang, D.}, \bibinfo{author}{Zhang, D.}, \bibinfo{author}{Han, J.}, \bibinfo{year}{2024}.
\newblock \bibinfo{title}{Deep generative adversarial reinforcement learning for semi-supervised segmentation of low-contrast and small objects in medical images}.
\newblock \bibinfo{journal}{IEEE Transactions on Medical Imaging} \bibinfo{volume}{43}, \bibinfo{pages}{3072--3084}.
%Type = Article
\bibitem[{Xu et~al.(2022)Xu, Zhou, Jin, Xie, Chen, Hu and He}]{xu2022framework}
\bibinfo{author}{Xu, Y.}, \bibinfo{author}{Zhou, B.}, \bibinfo{author}{Jin, S.}, \bibinfo{author}{Xie, X.}, \bibinfo{author}{Chen, Z.}, \bibinfo{author}{Hu, S.}, \bibinfo{author}{He, N.}, \bibinfo{year}{2022}.
\newblock \bibinfo{title}{A framework for urban land use classification by integrating the spatial context of points of interest and graph convolutional neural network method}.
\newblock \bibinfo{journal}{Computers, Environment and Urban Systems} \bibinfo{volume}{95}, \bibinfo{pages}{101807}.
%Type = Article
\bibitem[{Yan et~al.(2019)Yan, Ai, Yang and Yin}]{yan2019graph}
\bibinfo{author}{Yan, X.}, \bibinfo{author}{Ai, T.}, \bibinfo{author}{Yang, M.}, \bibinfo{author}{Yin, H.}, \bibinfo{year}{2019}.
\newblock \bibinfo{title}{A graph convolutional neural network for classification of building patterns using spatial vector data}.
\newblock \bibinfo{journal}{ISPRS journal of photogrammetry and remote sensing} \bibinfo{volume}{150}, \bibinfo{pages}{259--273}.
%Type = Article
\bibitem[{Yang et~al.(2022)Yang, Song, King and Xu}]{yang2022survey}
\bibinfo{author}{Yang, X.}, \bibinfo{author}{Song, Z.}, \bibinfo{author}{King, I.}, \bibinfo{author}{Xu, Z.}, \bibinfo{year}{2022}.
\newblock \bibinfo{title}{A survey on deep semi-supervised learning}.
\newblock \bibinfo{journal}{IEEE transactions on knowledge and data engineering} \bibinfo{volume}{35}, \bibinfo{pages}{8934--8954}.
%Type = Article
\bibitem[{Yao et~al.(2023)Yao, Zhu, Guo, Huang, Zhang, Yan, Dong, Jiang, Liu and Guan}]{yao2023unsupervised}
\bibinfo{author}{Yao, Y.}, \bibinfo{author}{Zhu, Q.}, \bibinfo{author}{Guo, Z.}, \bibinfo{author}{Huang, W.}, \bibinfo{author}{Zhang, Y.}, \bibinfo{author}{Yan, X.}, \bibinfo{author}{Dong, A.}, \bibinfo{author}{Jiang, Z.}, \bibinfo{author}{Liu, H.}, \bibinfo{author}{Guan, Q.}, \bibinfo{year}{2023}.
\newblock \bibinfo{title}{Unsupervised land-use change detection using multi-temporal poi embedding}.
\newblock \bibinfo{journal}{International Journal of Geographical Information Science} \bibinfo{volume}{37}, \bibinfo{pages}{2392--2415}.
%Type = Inproceedings
\bibitem[{Yao et~al.(2018)Yao, Fu, Liu, Hu and Xiong}]{yao2018representing}
\bibinfo{author}{Yao, Z.}, \bibinfo{author}{Fu, Y.}, \bibinfo{author}{Liu, B.}, \bibinfo{author}{Hu, W.}, \bibinfo{author}{Xiong, H.}, \bibinfo{year}{2018}.
\newblock \bibinfo{title}{Representing urban functions through zone embedding with human mobility patterns}, in: \bibinfo{booktitle}{Proceedings of the 27th International Joint Conference on Artificial Intelligence}, \bibinfo{publisher}{AAAI Press}. p. \bibinfo{pages}{3919–3925}.
%Type = Article
\bibitem[{Yong and Zhou(2024)}]{yong2024musecl}
\bibinfo{author}{Yong, X.}, \bibinfo{author}{Zhou, X.}, \bibinfo{year}{2024}.
\newblock \bibinfo{title}{Musecl: Predicting urban socioeconomic indicators via multi-semantic contrastive learning}.
\newblock \bibinfo{journal}{arXiv preprint arXiv:2407.09523} .
%Type = Article
\bibitem[{Yoon(2021)}]{yoon2021forecasting}
\bibinfo{author}{Yoon, J.}, \bibinfo{year}{2021}.
\newblock \bibinfo{title}{Forecasting of real gdp growth using machine learning models: Gradient boosting and random forest approach}.
\newblock \bibinfo{journal}{Computational Economics} \bibinfo{volume}{57}, \bibinfo{pages}{247--265}.
%Type = Article
\bibitem[{Yu et~al.(2017)Yu, Yin and Zhu}]{yu2017spatio}
\bibinfo{author}{Yu, B.}, \bibinfo{author}{Yin, H.}, \bibinfo{author}{Zhu, Z.}, \bibinfo{year}{2017}.
\newblock \bibinfo{title}{Spatio-temporal graph convolutional networks: A deep learning framework for traffic forecasting}.
\newblock \bibinfo{journal}{arXiv preprint arXiv:1709.04875} .
%Type = Inproceedings
\bibitem[{Zhang et~al.(2021)Zhang, Li, Li and Hui}]{zhang2021multi}
\bibinfo{author}{Zhang, M.}, \bibinfo{author}{Li, T.}, \bibinfo{author}{Li, Y.}, \bibinfo{author}{Hui, P.}, \bibinfo{year}{2021}.
\newblock \bibinfo{title}{Multi-view joint graph representation learning for urban region embedding}, in: \bibinfo{booktitle}{Proceedings of the twenty-ninth international conference on international joint conferences on artificial intelligence}, pp. \bibinfo{pages}{4431--4437}.
%Type = Article
\bibitem[{Zhang et~al.(2017)Zhang, Du and Wang}]{zhang2017hierarchical}
\bibinfo{author}{Zhang, X.}, \bibinfo{author}{Du, S.}, \bibinfo{author}{Wang, Q.}, \bibinfo{year}{2017}.
\newblock \bibinfo{title}{Hierarchical semantic cognition for urban functional zones with vhr satellite images and poi data}.
\newblock \bibinfo{journal}{ISPRS Journal of Photogrammetry and Remote Sensing} \bibinfo{volume}{132}, \bibinfo{pages}{170--184}.

\end{thebibliography}

%%       \citet{<label>} ==> Lamport (1994)
%%       \citep{<label>} ==> (Lamport, 1994)
%%



\end{document}